\documentclass{article}

\usepackage{microtype}
\usepackage{graphicx}
\usepackage{subcaption}
\usepackage{booktabs} 

\usepackage{hyperref}

\usepackage[accepted]{icml2026}

\usepackage{amsmath}
\usepackage{amssymb}
\usepackage{mathtools}
\usepackage{amsthm}

\usepackage{wrapfig}
\usepackage{comment}
\usepackage{nicefrac}       
\usepackage{microtype}      
\usepackage{xcolor}         
\usepackage{colortbl} 
\usepackage{subcaption}
\usepackage{multirow}
\usepackage{verbatim}
\usepackage{diagbox}
\usepackage{bm}
\usepackage{makecell}
\usepackage{graphicx}
\usepackage{amsmath}
\usepackage[english]{babel}
\usepackage{mathtools}
\usepackage[utf8]{inputenc}
\usepackage{csquotes}
\usepackage{enumitem}
\usepackage{soul}
\usepackage{lipsum} 
\usepackage{dsfont}

\usepackage{booktabs} 
\usepackage{caption} 
\usepackage{rotating} 
\usepackage{multirow} 
\usepackage{lipsum} 
\usepackage{colortbl}
\usepackage{arydshln}
\usepackage{amsthm}
\usepackage{amssymb}
\usepackage{bbm}
\usepackage[table]{xcolor}
\newcommand{\rowhighlight}{\rowcolor{deltaBg}}
\definecolor{deltaBg}{RGB}{235, 250, 240} 
\definecolor{darkred}{RGB}{177, 38, 26}
\definecolor{darkblue}{RGB}{67, 116, 177}
\definecolor{darkgreen}{rgb}{0.0, 0.5, 0.0}

\usepackage[most]{tcolorbox}
\usepackage{longtable}
\usepackage{booktabs}
\usepackage{titletoc}
\usepackage{xcolor}

\usepackage[capitalize,noabbrev]{cleveref}

\newtcolorbox{promptbox}[3][Judge Prompt]{%
  enhanced,
  breakable,
  colback=black!5!white,
  arc=5pt,
  boxrule=0.5pt,
  fonttitle=\bfseries,
  title=#1,
  before upper={\small},
  fontupper=\fontfamily{ptm}\selectfont,
  colframe=#2,                 
  colbacktitle=#2!70!black,    
  coltitle=white,             
  attach boxed title to top left={xshift=2mm, yshift*=-1mm},
  boxed title style={
    sharp corners,
    boxrule=0pt,
    arc=3pt,
    left=4pt,right=4pt,top=2pt,bottom=2pt
  },
  label=#3,
}
\definecolor{lightorange}{RGB}{253, 208, 162}
\definecolor{mintgreen}{RGB}{173, 223, 173}

\theoremstyle{plain}

\theoremstyle{definition}

\theoremstyle{remark}

\usepackage[textsize=tiny]{todonotes}

\icmltitlerunning{Decomposed On-Policy Distillation for Vision-Language Reasoning: Steering Gradients for Visual Grounding}

\begin{document}

\twocolumn[
  \icmltitle{Decomposed On-Policy Distillation for Vision-Language Reasoning: Steering Gradients for Visual Grounding}



  \icmlsetsymbol{equal}{*}

\begin{icmlauthorlist}
\icmlauthor{Hee Suk Yoon}{yyy}
\icmlauthor{Eunseop Yoon}{yyy}
\icmlauthor{Jaehyun Jang}{yyy}
\icmlauthor{SooHwan Eom}{yyy}
\icmlauthor{Ji Woo Hong}{yyy}\\
\icmlauthor{Mark Hasegawa-Johnson}{xxx}
\icmlauthor{Qi Dai}{zzz}
\icmlauthor{Chong Luo}{zzz}
\icmlauthor{Chang D. Yoo}{yyy}
\end{icmlauthorlist}

  \icmlaffiliation{yyy}{Korea Advanced Institute of Science and Technology (KAIST),}
  \icmlaffiliation{xxx}{University of Illinois Urbana-Champaign (UIUC),}
  \icmlaffiliation{zzz}{Microsoft Research Asia (MSRA)}

  \icmlcorrespondingauthor{Chang D. Yoo}{cd\_yoo@kaist.ac.kr}

  \icmlkeywords{Machine Learning, ICML}

  \vskip 0.3in
]



\printAffiliationsAndNotice{}  
\begin{abstract}
While on-policy distillation offers dense supervision for training small reasoning models, its optimization dynamics in the multimodal domain remain under-explored. In this work, we challenge the standard monolithic view of Vision-Language Model (VLM) distillation by mathematically decomposing the loss into two distinct components: the language prior and visual grounding. Our analysis uncovers that gradient vectors for these components are nearly orthogonal, indicating that the objective of aligning with the teacher's language distribution is geometrically independent from the objective of matching its visual perception. Consequently, standard optimization passively follows a suboptimal compromise trajectory that implicitly balances the two objectives. Hypothesizing that visual grounding constitutes the primary bottleneck for vision-language reasoning, we introduce \textbf{Visual Gradient Steering (VGS)}, a method that dynamically reorients the update vector to prioritize the visual subspace. Experimental results on multiple distillation settings and complex multimodal benchmarks demonstrate that VGS significantly outperforms the standard monolithic formulation of on-policy distillation, achieving superior grounding with minimal training overhead. The code is publicly accessible at \href{https://github.com/hee-suk-yoon/Decomposed_OPD}{https://github.com/hee-suk-yoon/Decomposed\_OPD}.
\end{abstract}
\vspace{-15pt}
\section{Introduction}

The recent surge in reasoning models has demonstrated that scaling inference-time compute by enabling models to generate intermediate reasoning steps can dramatically enhance problem-solving capabilities~\citep{gemini2.5, yang2025qwen3, qwq32b}. To train these reasoning models, Reinforcement Learning with Verifiable Rewards (RLVR) has emerged as the standard for domains like mathematics and coding where ground truth is easily verifiable~\citep{qwen2_5_math, guo2025deepseek, grpo, yoon2025pacr, yoon2026pdcr}. However, for smaller models, RLVR encounters a cold start problem. Sparse, outcome-based rewards fail to provide sufficient signal for models that initially lack robust reasoning policies.

\begin{figure}[t]
	\centering
	\includegraphics[width=1.0\linewidth]{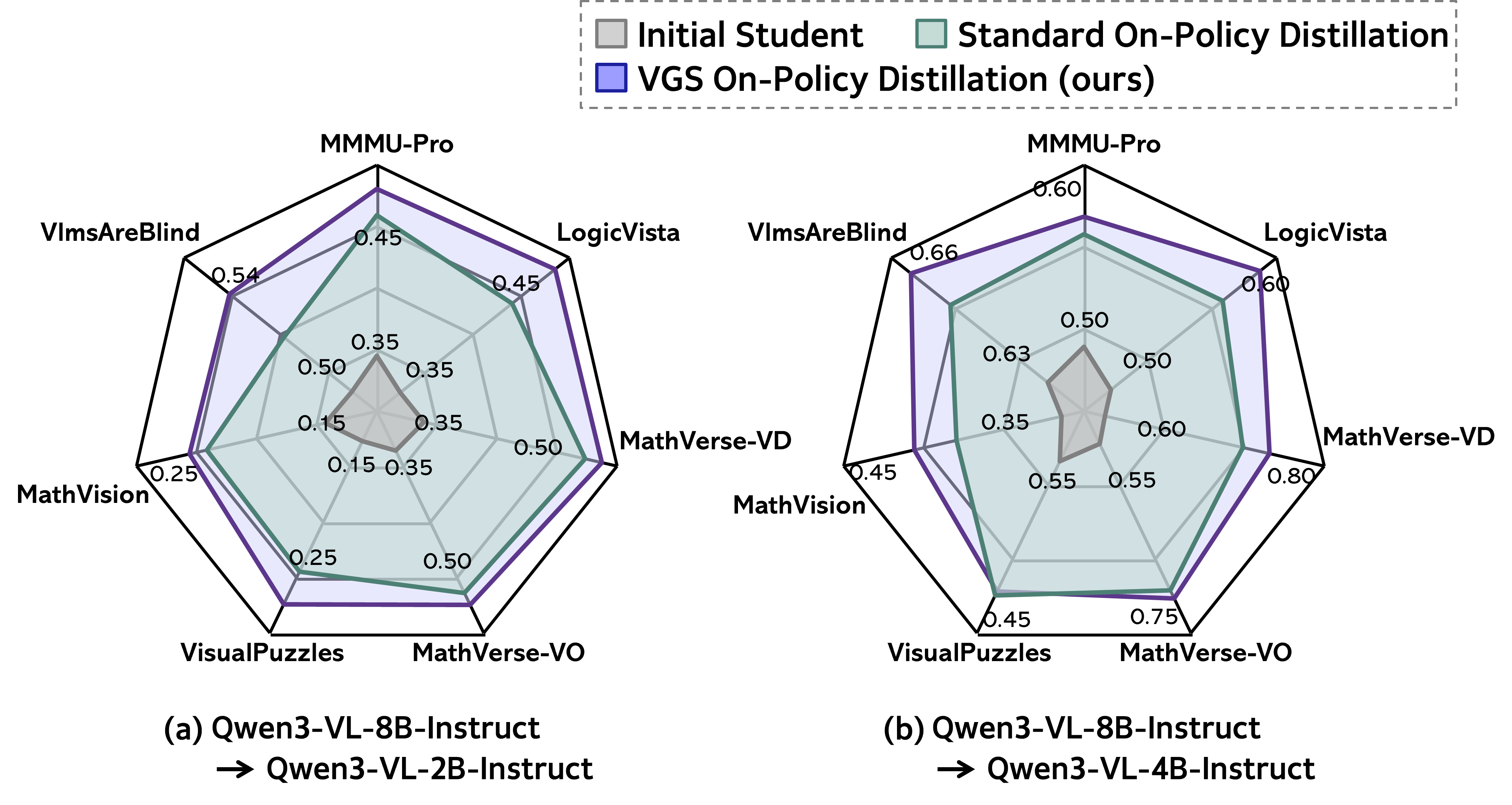}
\caption{\textbf{Visual Gradient Steering (VGS) outperforms standard monolithic distillation.} We compare the reasoning performance of student models distilled from a 8B teacher. VGS (purple) consistently surpasses the standard baseline (green) across diverse multimodal benchmarks for both \textbf{(a)} 2B and \textbf{(b)} 4B students, demonstrating superior visual grounding.}
    \label{fig: intro_example}
\end{figure}

\begin{figure*}[t!]
\centering
    \includegraphics[width=1.00\textwidth]{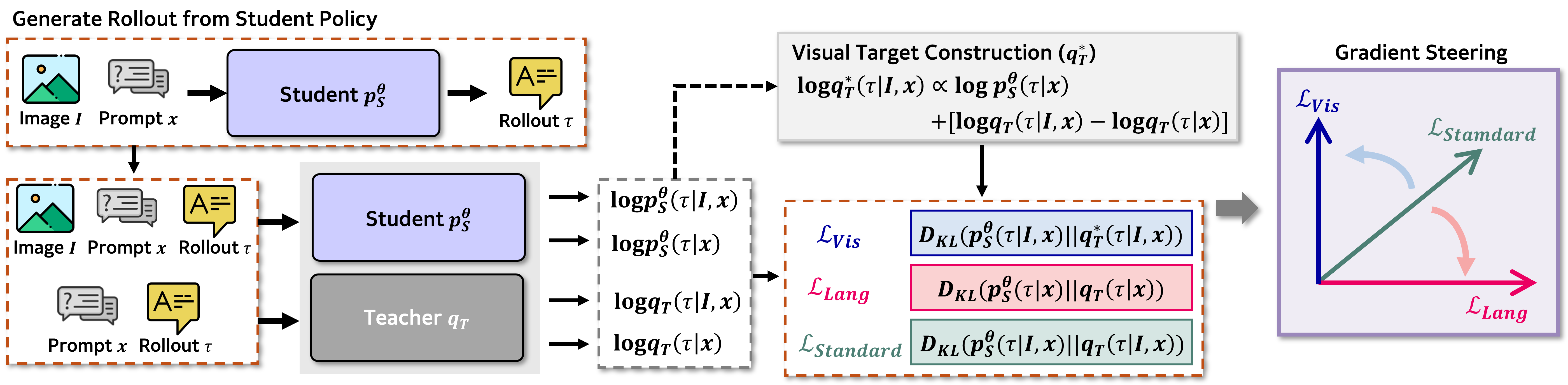}
\caption{\textbf{Overview of Visual Gradient Steering (VGS).} 
\textbf{(Left)} Given a multimodal query (Image $I$ and Prompt $x$), we sample a rollout $\tau$ from the student policy and compute log-probabilities from both the student and teacher models under multimodal ($I, x$) and unimodal ($x$, text-only) contexts. 
\textbf{(Middle)} We decompose the standard monolithic objective into two distinct components: a \textit{Language Prior} ($\mathcal{L}_{\text{Lang}}$) that matches the teacher's text-only distribution, and a \textit{Visual Grounding} objective ($\mathcal{L}_{\text{Vis}}$) that targets a constructed distribution $q_T^*$ isolating the teacher's visual information gain. 
\textbf{(Right)} Our geometric analysis reveals that standard distillation ($\mathcal{L}_{\text{Standard}}$) acts as a passive compromise between these often orthogonal objectives. VGS explicitly steers the gradient update toward the visual subspace ($\mathcal{L}_{\text{Vis}}$), prioritizing perceptual fidelity over generic language modeling.}
\label{fig:overview}
    \label{fig: method}
\end{figure*}

On-Policy Distillation~\citep{gkd, xi2024training, srl} offers a superior alternative. Unlike RLVR which relies on delayed outcome feedback, on-policy distillation leverages a stronger teacher model to provide dense, token-level supervision on the student’s \textit{own} rollouts. This overcomes the reward sparsity problem while still allowing the student to learn error recovery and reasoning coherence, akin to a chess coach critiquing every move rather than just the final result.

Despite its success in text-only domains, on-policy distillation remains under-explored in Vision-Language Models (VLMs). For instance, recent models such as Qwen3-VL~\citep{bai2025qwen3vltechnicalreport} explicitly restrict distillation to text-only data to fine-tune the LLM backbone, neglecting visual grounding alignment.

We find that a direct extension of standard on-policy distillation to the multimodal domain typically defaults to optimizing a single, monolithic objective. In this work, we challenge this formulation by mathematically decomposing the loss into distinct \textit{Language Prior} and \textit{Visual Grounding} components, uncovering a fundamental tension between them (Figure \ref{fig: method}). Through gradient analysis, we identify a geometric pathology where the gradient vectors for these components are frequently orthogonal. This indicates that the objective of aligning with the teacher's language distribution is geometrically independent from the objective of matching its visual perception. Consequently, standard distillation naively sums these gradients and passively follows a suboptimal compromise trajectory that fails to strictly enforce perceptual fidelity.

Hypothesizing that visual grounding constitutes the primary bottleneck for vision-language reasoning, we introduce \textbf{Visual Gradient Steering (VGS)}. This gradient update rule explicitly prioritizes the visual subspace by dynamically reorienting the optimization trajectory to maximize the alignment with the teacher's visual information gain. By ensuring that the model concentrates its update budget on resolving perceptual ambiguities rather than generic language modeling, VGS achieves superior alignment performance (Figure \ref{fig: intro_example}). \textbf{In summary, our key contributions are:}
\begin{itemize}[left=0em]
\setlength\itemsep{0em}
    \item \textbf{Geometric Analysis of VLM Distillation.} We provide the first gradient-level analysis of on-policy distillation in VLMs, revealing the orthogonality between language and vision gradients and identifying the suboptimal "compromise trajectory" of standard monolithic formulations.
    \item \textbf{Visual Gradient Steering (VGS).} We propose a novel optimization objective that separates visual and language supervision, steering the gradient update to strictly enforce visual grounding without degrading linguistic fluency.
    \item \textbf{Superior Reasoning Alignment.} We show that VGS significantly outperforms the standard monolithic formulation of on-policy distillation on complex multimodal reasoning benchmarks, with minimal training overhead. 
\end{itemize}

\section{Background and Problem Formulation}
\label{sec:background}

\subsection{Distillation for Autoregressive Reasoning}
\label{sec:distillation_basics}

In reasoning tasks, given a query $x$, the model generates a multi-step reasoning trajectory $\tau$. Formally, $\tau = (h_1, \dots, h_T, \hat{y})$ consists of a sequence of intermediate reasoning steps $\{h_k\}_{k=1}^T$, commonly termed a Chain-of-Thought (CoT), followed by a final predicted answer $\hat{y}$.

In the context of distillation, we optimize a student model $p_{\text{S}}^\theta$ to match a fixed teacher model $q_{\text{T}}$ using dense, token-level supervision. We quantify the discrepancy between the teacher and student for a trajectory $\tau$ using the sequence-level Kullback-Leibler (KL) divergence, defined as the average token-level Forward KL divergence across the sequence: 
\[
\ell_{\text{Forward}}(\tau) \triangleq \frac{1}{|\tau|} \sum_{t=1}^{|\tau|} D_{\text{KL}}\big( q_\text{T}(\cdot \mid \tau_{<t}, x) \parallel p_\text{S}^\theta(\cdot \mid \tau_{<t}, x) \big).
\]
The notations $q_T(\cdot \mid \tau_{<t}, x)$ and $p_S^\theta(\cdot \mid \tau_{<t}, x)$ denote the \textbf{full categorical distributions} over the vocabulary $\mathcal{V}$ at step $t$ for the teacher and student, respectively.

Standard \textit{Sequence-Level Knowledge Distillation (SeqKD)} \cite{kim2016sequencelevelknowledgedistillation, alpaca}, often referred to as \textbf{Off-Policy Distillation}, minimizes the \textbf{Forward KL Divergence}. This objective computes the expected sequence divergence over trajectories sampled strictly from the teacher: 
\begin{equation}
\label{eq:forward_kl}
\mathcal{L}_{\text{Off-Policy}} = \mathop{\mathbb{E}}\limits_{\tau \sim q_\text{T}} \left[ \ell_{\text{Forward}}(\tau) \right].
\end{equation}
While effective for initialization, Forward KL is mode-covering, where it forces the student to assign probability mass to all modes of the teacher's distribution. For reasoning tasks, this is often suboptimal, since if the student lacks the capacity to model the full complexity of the teacher, it forces the student to bridge distinct modes with low-probability transitions, leading to hallucinations and implausible reasoning chains~\cite{gu2024minillm, gkd}. Furthermore, this off-policy approach suffers from \textit{exposure bias}: the student is never trained to recover from its own autoregressive errors.

To address these limitations, recent works \cite{gkd, gu2024minillm} advocate for \textbf{On-Policy Distillation} via the \textbf{Reverse KL Divergence}. We first define the trajectory-wise Reverse KL for a trajectory $\tau$ as:
\[
\ell_{\text{Reverse}}(\tau) \triangleq \frac{1}{|\tau|} \sum_{t=1}^{|\tau|} D_{\text{KL}}\big( p_\text{S}^\theta(\cdot \mid \tau_{<t}, x) \parallel q_\text{T}(\cdot \mid \tau_{<t}, x) \big).
\]
This approach minimizes the expected sequence-level divergence over trajectories sampled from the student:
\begin{equation}
\label{eq:on_policy_obj}
\mathcal{L}_{\text{On-Policy}} = \mathop{\mathbb{E}}\limits_{\tau \sim p_S^\theta} \left[ \ell_{\text{Reverse}}(\tau) \right].
\end{equation}
This formulation offers two primary benefits. First, Reverse KL is \textbf{mode-seeking}; it penalizes the student for generating samples unlikely under the teacher ($q_T \approx 0$), prioritizing high-confidence reasoning paths over broad coverage~\cite{gu2024minillm, gkd}. Second, by sampling $\tau \sim p_S^\theta$, the student learns from its own rollouts (on-policy), effectively closing the training-inference gap.

\subsection{The Standard Monolithic Multimodal Objective}
\label{sec:problem_formulation}
In the Vision-Language domain, we extend this on-policy formulation to include visual context. Given an input image $I$ and text prompt $x$, we define the \textbf{monolithic trajectory-wise divergence} for a trajectory $\tau$ as:
\begin{equation}
\label{eq:mono_pointwise}
\small
\ell_{\text{Standard}}(\tau) \triangleq \frac{1}{|\tau|} \sum_{t=1}^{|\tau|} D_{\text{KL}}\big( p_\text{S}^\theta(\cdot \mid \tau_{<t}, I, x) \parallel q_\text{T}(\cdot \mid \tau_{<t}, I, x) \big).
\end{equation}
Standard approaches optimize the expectation of this quantity over trajectories sampled from the student:
\begin{equation}
\label{eq:monolithic}
\mathcal{L}_{\text{Standard}} = \mathop{\mathbb{E}}\limits_{\tau \sim p_S^\theta(\cdot \mid I, x)} \left[ \ell_{\text{Standard}}(\tau) \right].
\end{equation}
However, as we demonstrate next, this treats the learning signal as a unified whole, obscuring the distinct contributions of \textit{Language Prior Matching} and \textit{Visual Grounding}.

\begin{figure*}[ht!]
\centering
    \includegraphics[width=1.0\textwidth]{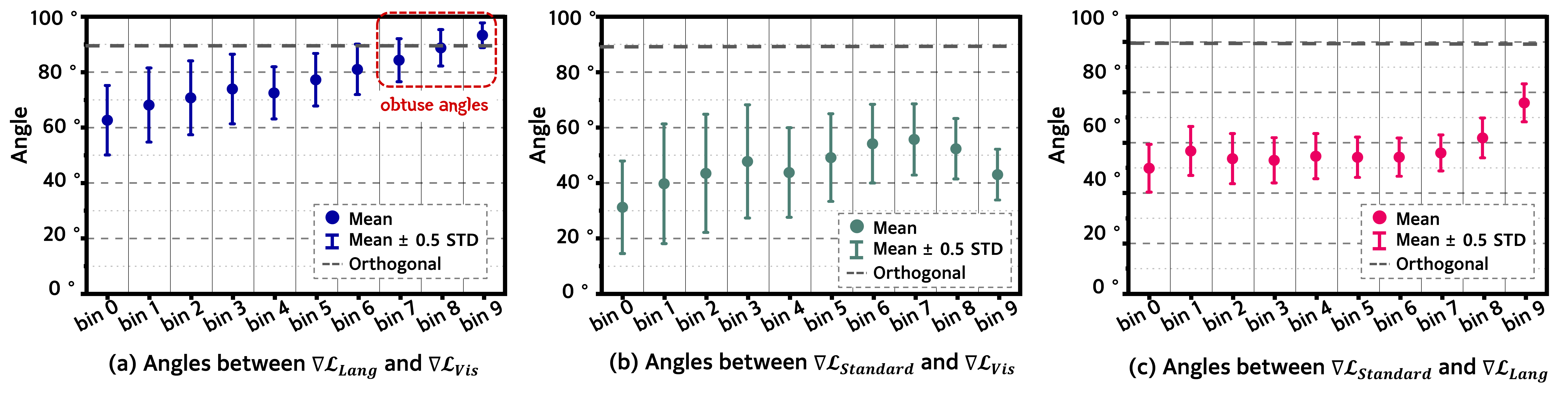}
\caption{\textbf{Geometric Analysis of Gradient Dynamics.} 
\textbf{(a)} As visual dependency increases, the angle between Language and Visual gradients widens, approaching \textbf{orthogonality} in high-dependency regions. The emergence of obtuse angles ($>90^\circ$, dashed box) at the extreme specifically motivates the \textit{Language Preservation} regularizer (Eq.~\ref{eq:lp_term}) to prevent gradient conflict. 
\textbf{(b, c)} The standard monolithic gradient ($\nabla\mathcal{L}_{\text{Standard}}$) acts as a passive bisector, maintaining a relatively static compromise orientation ($\approx 40^\circ-50^\circ$) relative to both components, failing to fully align with the visual signal even when perceptual necessity is maximal (Bin 9).}
    \label{fig: grad_angle_analysis}
\end{figure*}

\section{Revisiting On-Policy Distillation in Vision-Language}
\label{sec:gradient_analysis}

\subsection{Decomposing the Monolithic Objective}
\label{sec:decomposition}
By applying Bayes' Rule, we can factorize the conditional probability of any multimodal generator $p$ to isolate the underlying components of the objective. The log-likelihood decomposes into a language prior, a visual likelihood, and a sequence-independent constant:
\begin{equation}
\label{eq:bayes_decomposition}
\small
\log p(\tau \mid I, x) =
\underbrace{\log p(\tau \mid x)}_{\text{Language Prior}} + \underbrace{\log p(I \mid \tau, x)}_{\text{Visual Likelihood}} - \log p(I \mid x).
\end{equation}
Applying this identity to both the student $p_S^\theta$ and teacher $q_T$ reveals that the standard monolithic objective implicitly combines two distinct goals: matching the \textit{Language Prior} (reasoning style) and matching the \textit{Visual Likelihood} (perceptual grounding). Note that the evidence term $\log p(I \mid x)$ is constant with respect to the generated sequence $\tau$ and acts solely as a normalizing factor.

\paragraph{I. Language Prior Matching ($\mathcal{L}_{\text{Lang}}$).}
This objective aligns the student's unimodal distribution $p_S^\theta(\cdot \mid x)$ with the teacher's $q_T(\cdot \mid x)$, ensuring that the base reasoning style is transferred independent of visual context $I$. We first define the \textbf{trajectory-wise language divergence}:
\begin{equation}
\label{eq:lang_pointwise}
\small
\ell_{\text{Lang}}(\tau) \triangleq \frac{1}{|\tau|} \sum_{t=1}^{|\tau|} D_{KL}\big( p_\text{S}^\theta(\cdot \mid \tau_{<t}, x) \parallel q_\text{T}(\cdot \mid \tau_{<t}, x) \big).
\end{equation}
We minimize the expectation of this divergence. Crucially, while we compute gradients on the unimodal distributions, we sample trajectories $\tau$ from the full multimodal policy $p_S^\theta(\cdot \mid I, x)$ to maintain on-policy alignment with the student's actual generation distribution:
\begin{equation}
\label{eq:lang_loss}
\mathcal{L}_{\text{Lang}} = \mathop{\mathbb{E}}\limits_{\tau \sim p_S^\theta(\cdot \mid I, x)} \left[ \ell_{\text{Lang}}(\tau) \right].
\end{equation}
\paragraph{II. Visual Grounding Matching ($\mathcal{L}_{\text{Vis}}$).} This objective isolates the \textit{Visual Likelihood} term identified in Eq.~\ref{eq:bayes_decomposition}. This term, $\log p(I \mid \tau, x)$, represents the model's \textbf{perceptual sensitivity}, quantifying how well the generated reasoning causally explains the visual input. Although explicit computation of the likelihood is intractable in autoregressive models, Bayes' Rule reveals that it differs from the computable \textbf{Visual Information Gain}---the log-ratio of the posterior to the prior---only by a sequence-independent constant:
\begin{equation}
\small
\log p(I \mid \tau, x) =
\underbrace{\left[ \log p(\tau \mid I, x) - \log p(\tau \mid x) \right]}_{\text{Visual Information Gain}} + \log p(I \mid x).
\end{equation}
Since the evidence term $\log p(I \mid x)$ depends only on the image and is constant with respect to the trajectory $\tau$, matching the teacher's information gain is mathematically equivalent to aligning the student's visual perception.

Formally, we achieve this by constructing a \textbf{visual target distribution} $q_T^*$ that retains the student's language prior but substitutes its visual likelihood with that of the teacher:
\begin{equation}
\label{eq:hybrid_definition}
q^*_T(\tau \mid I, x) \propto p_S^\theta(\tau \mid x) \cdot q_T(I \mid \tau, x).
\end{equation}
Substituting the Bayesian expansion of the teacher's likelihood into Eq.~\ref{eq:hybrid_definition} yields the tractable target logits in log-space:
\begin{equation}
\label{eq:target_q}
\begin{split}
\log q^*_T(\tau & \mid I, x) = \log p_S^\theta(\tau \mid x) + \\
& (\log q_T(\tau \mid I, x) - \log q_T(\tau \mid x)) - \log Z^*.
\end{split}
\end{equation}
Crucially, the marginal evidence term $\log q_T(I \mid x)$ acts as a scalar offset across all tokens. In our autoregressive formulation, this \textit{constant is absorbed into the local partition function} $Z^*$, which is computed implicitly via the Softmax operator at each decoding step.

We finally define the \textbf{trajectory-wise visual divergence} $\ell_{\text{Vis}}(\tau)$ against this target and its corresponding expectation:
\begin{align}
\small
\ell_{\text{Vis}}(\tau) & \triangleq \frac{1}{|\tau|} \sum_{t=1}^{|\tau|} D_{KL}\big(p_S^\theta(\cdot \mid \tau_{<t}, I, x) \parallel q^*_T(\cdot \mid \tau_{<t}, I, x)\big), \nonumber \\
\mathcal{L}_{\text{Vis}} & = \mathop{\mathbb{E}}\limits_{\tau \sim p_S^\theta(\cdot \mid I, x)} \left[ \ell_{\text{Vis}}(\tau) \right]. \label{eq:visual_loss_final}
\end{align}

\subsection{Geometric Analysis of Gradient Dynamics}
\label{sec:geometric_analysis}

To understand the optimization dynamics of the standard monolithic objective, we empirically analyzed the geometric relationships between its decomposed gradient vectors. Specifically, we investigate how the standard update $\nabla \mathcal{L}_{\text{Standard}}$ positions itself relative to the \textit{Language Prior} ($\nabla \mathcal{L}_{\text{Lang}}$) and \textit{Visual Grounding} ($\nabla \mathcal{L}_{\text{Vis}}$) gradients across tokens with varying degrees of visual dependency.

\textbf{Analysis Setup.} We analyze the gradient dynamics on validation prompts, using trajectories sampled on-policy from the full multimodal student policy, $\tau \sim p_S^\theta(\cdot \mid I, x)$. To categorize the perceptual necessity of each generated token, we define the \textbf{Visual Dependency Score (VDS)} using the teacher's distribution:
\begin{equation}
\label{eq: VDS}
\text{VDS}_t = D_{KL}(q_T(\cdot \mid \tau_{<t}, I, x) \parallel q_T(\cdot \mid \tau_{<t}, x)).
\end{equation}
We bin all generated tokens into 10 equal-frequency quantiles based on their VDS; \textbf{bin 0} represents tokens with minimal visual dependence, while \textbf{bin 9} represents tokens where the teacher relies heavily on visual evidence.

Figure~\ref{fig: grad_angle_analysis} visualizes the angular relationships across these bins. Our analysis highlights two key geometric properties:

\begin{figure*}[ht!]
\centering
    \includegraphics[width=1.00\textwidth]{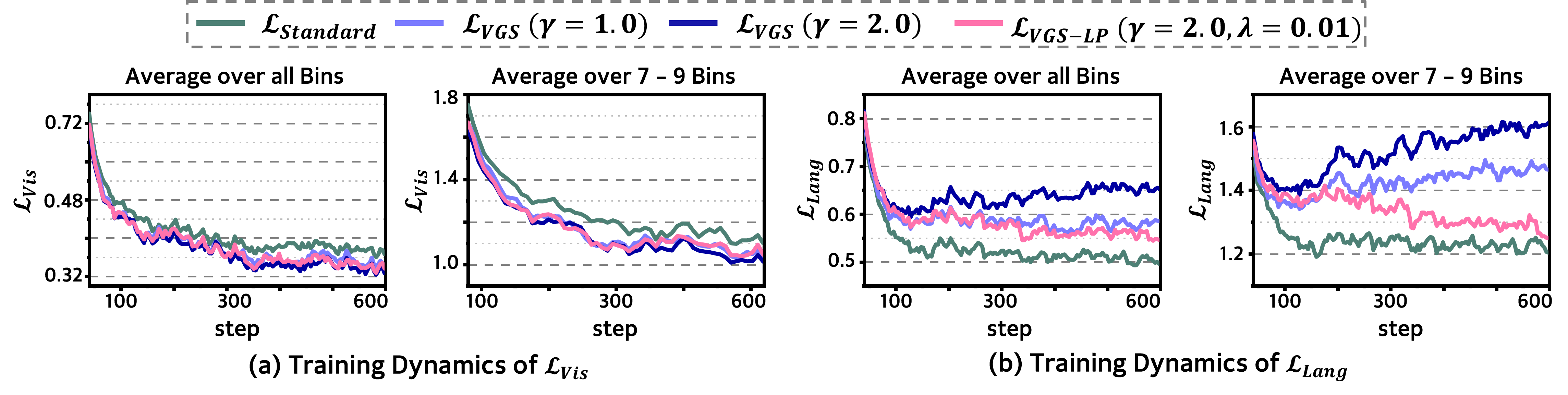}
\caption{\textbf{Training Dynamics.} We compare the evolution of decomposed loss components during training across different methods.
\textbf{(a) Visual Grounding ($\mathcal{L}_{\text{Vis}}$):} VGS (blue) significantly accelerates visual learning compared to the standard monolithic baseline ($\mathcal{L}_{\text{Standard}}$, green), especially for high-dependency tokens (Bins 7--9).
\textbf{(b) Language Prior ($\mathcal{L}_{\text{Lang}}$):} Without regularization, aggressive visual steering ($\gamma=2.0$, dark blue) causes the language prior to diverge (``unlearning''), particularly in high-conflict visual regimes (Bins 7--9). Our final method with Language Preservation ($\mathcal{L}_{\text{VGS-LP}}$, pink) successfully prevents this degradation while maintaining superior visual grounding.}
    \label{fig: training loss curve}
\end{figure*}

\paragraph{I. Orthogonality at the Visual Extremes (Figure~\ref{fig: grad_angle_analysis}-(a)).} We first examine the relationship between the language and visual gradients. We observe a distinct monotonic trend: as the visual dependency of the token increases, the angle between $\nabla \mathcal{L}_{\text{Lang}}$ and $\nabla \mathcal{L}_{\text{Vis}}$ widens.
\begin{itemize}
    \item In low-dependency regions (bin 0), the gradients are moderately aligned ($\theta \approx 60^\circ$).
    \item \textbf{In high-dependency regions (bin 9), the gradients become nearly orthogonal ($\theta \approx 92^\circ$).}
\end{itemize}

\paragraph{II. The Standard Loss as a Static Bisector (Figure~\ref{fig: grad_angle_analysis}-(b), (c)).} Next, we analyze the orientation of the standard monolithic gradient $\nabla \mathcal{L}_{\text{Standard}}$. Figures~\ref{fig: grad_angle_analysis}-(b) and (c) show the angle of the standard gradient relative to the Visual and Language components, respectively.

We observe that the standard objective consistently maintains a ``compromise'' trajectory. Even for the most visually critical tokens (Bin 9), where the ideal update should arguably align with the visual evidence, $\nabla \mathcal{L}_{\text{Standard}}$ retains a significant offset of $\approx 42^\circ$ from $\nabla \mathcal{L}_{\text{Vis}}$. Simultaneously, it remains similarly distant from the language gradient ($\approx 50^\circ$). This \textbf{geometric rigidity} indicates that the standard objective acts as a \textbf{passive bisector}: it creates a fixed average between the language and visual directions, treating both signals as equally important. 
\paragraph{Hypothesis: Breaking the Optimization Symmetry.} We hypothesize that the standard objective is suboptimal because it enforces geometric symmetry on an inherently asymmetric task. Since performance is constrained by a \textbf{perceptual bottleneck}, and language gradients are orthogonal to visual improvements, the standard compromise update is inefficient. We propose that explicitly \textbf{steering} the gradient toward the visual subspace introduces a necessary inductive bias, prioritizing the resolution of perceptual ambiguities over generic language modeling.

\section{Method: Visual Gradient Steering}
\label{sec:method}
Motivated by the geometric pathology observed in Sec.~\ref{sec:geometric_analysis}, we propose \textbf{Visual Gradient Steering (VGS)} to construct a gradient update that prioritizes the visual subspace.

\paragraph{The Steered Objective.} We first define the trajectory-wise steered objective $\ell_{\text{VGS}}(\tau)$, which augments the standard distillation loss $\ell_{\text{Standard}}(\tau)$ with an auxiliary visual term:
\begin{equation}
\label{eq:vgs_pointwise}
\ell_{\text{VGS}}(\tau) \triangleq \ell_{\text{Standard}}(\tau) + \gamma \ell_{\text{Vis}}(\tau),
\end{equation}
where $\ell_{\text{Vis}}(\tau)$ corresponds to the visual grounding objective derived in Eq.~\ref{eq:visual_loss_final}, and $\gamma \ge 0$ is the \textit{steering coefficient} that controls the strength of the visual correction. The final loss function is then minimizing the expected steered objective:
\begin{equation}
\label{eq:vgs_final}
\mathcal{L}_{\text{VGS}} = \eta_{\text{VGS}}(\gamma) \cdot \mathop{\mathbb{E}}\limits_{\tau \sim p_S^\theta(\cdot \mid I, x)} \left[ \ell_{\text{VGS}}(\tau) \right].
\end{equation}
\paragraph{Gradient Norm Normalization.} To ensure the optimization remains stable, we introduce a scaling factor $\eta_{\text{VGS}}(\gamma)$. We define $\eta_{\text{VGS}}(\gamma)$ such that the norm of the steered gradient equals the norm of the original standard gradient:
\begin{equation}
\label{eq:beta_scaling}
\eta_{\text{VGS}}(\gamma) \triangleq \frac{\| \nabla_\theta \mathcal{L}_{\text{Standard}} \|_2}{\| \nabla_\theta \mathcal{L}_{\text{Standard}} + \gamma \nabla_\theta \mathcal{L}_{\text{Vis}} \|_2}.
\end{equation}
This normalization ensures that VGS changes only the \textit{direction} of the update—steering it toward the visual subspace—without altering the \textit{magnitude} of the parameter steps. This effectively decouples the ``steering" (controlled by $\gamma$) from the ``learning rate" (controlled by the optimizer), a stability principle widely adopted in multi-task learning~\cite{chen2018gradnorm}. In practice, to avoid the computational overhead of dynamic norm calculation, we set $\eta_{\text{VGS}}(\gamma)$ to a fixed constant dependent on $\gamma$, as empirical analysis in Appendix \ref{app:gradient_norm_eta} confirms the gradient magnitude ratio remains stable throughout training.

\begin{table*}[!t]
    \scriptsize
    \setlength{\extrarowheight}{0.5pt}
    \setlength{\tabcolsep}{6pt}
    \centering
    \vspace{-5pt}
\caption{\textbf{Main Results on Vision-Language Reasoning Benchmarks.} We compare the distillation performance of Visual Gradient Steering (VGS) against the standard monolithic baseline. All student models (2B and 4B) are distilled from the same Qwen3-VL-8B-Instruct teacher trained with GRPO. VGS consistently outperforms the standard approach across all benchmarks, achieving higher accuracy in both greedy decoding (Acc@1) and stochastic sampling (Acc@16).}
        \label{tab: main}
    \resizebox{1.0\textwidth}{!}{%
    \begin{tabular}{l|cc|cccc|cc}
    \toprule
        \multirow{2}{*}{\textbf{Benchmark}} & \textbf{Teacher} &  \textbf{\shortstack{Initial \\ Student}} & \multicolumn{2}{c}{\textbf{\shortstack{Standard  \\ On-Policy Distillation}}} & \multicolumn{2}{c|}{\textbf{\shortstack{VGS \\ On-Policy Distillation (ours)}}} & \multicolumn{2}{c}{\textbf{Improvement}} \\[1ex]
        \cdashline{2-9}
                  &  \rule{0pt}{5ex}  \shortstack{Acc@1 \\ (greedy)} & \shortstack{Acc@1 \\ (greedy)}  &  \shortstack{Acc@1 \\ (greedy)} & \shortstack{Acc@16 \\ (T=1.0)}  &  \shortstack{Acc@1 \\ (greedy)} & \shortstack{Acc@16 \\ (T=1.0)} & \shortstack{Acc@1 \\ (greedy)} & \shortstack{Acc@16 \\ (T=1.0)} \\
        \hline 
        \rowhighlight
        & \textbf{8B} & \textbf{2B} & \multicolumn{4}{c|}{\textbf{Qwen3-VL-8B-Instruct $\rightarrow$ Qwen3-VL-2B-Instruct}} & & \\
        \hline
        MMMU-Pro-4 & 62.03  & 34.51  & 45.83  & 47.33  & \textbf{48.07}  & \textbf{48.34} & \textbf{\color{darkgreen}+2.14}  & \textbf{\color{darkgreen}+1.01} \\
        LogicVista  &  60.01  & 36.83  & 45.53  & 46.19 & \textbf{48.88}  & \textbf{46.47} & \textbf{\color{darkgreen}+3.35} & \textbf{\color{darkgreen}+0.28} \\
        MathVerse-VD & 79.63  & 35.88  & 56.02 & 59.68 & \textbf{58.10}  & \textbf{60.10} & \textbf{\color{darkgreen}+2.08}  & \textbf{\color{darkgreen}+0.42} \\
        MathVerse-VO &  73.85  & 35.32  & 54.59 & 55.76 & \textbf{56.19}  & \textbf{56.52} & \textbf{\color{darkgreen}+1.61}  & \textbf{\color{darkgreen}+0.76} \\
        VisualPuzzles &  43.15 & 13.36  & 28.08 & 30.64 &  \textbf{31.76} & \textbf{31.59} & \textbf{\color{darkgreen}+3.68} & \textbf{\color{darkgreen}+0.95} \\
        MathVision &  44.14  & 14.28  & 24.14 & 25.37 &  \textbf{25.59} & \textbf{26.73} & \textbf{\color{darkgreen}+1.45} & \textbf{\color{darkgreen}+1.36} \\
        VlmsAreBlind &  66.79  & 49.03  & 51.86 & 50.52 &  \textbf{54.11} & \textbf{53.24} & \textbf{\color{darkgreen}+2.26} & \textbf{\color{darkgreen}+2.72} \\
        \hline
        \textbf{Average} &  61.37  & 31.32  &  43.74 & 45.07  & \textbf{46.10}  & \textbf{46.14} & \textbf{\color{darkgreen}+2.37}  & \textbf{\color{darkgreen}+1.07} \\
        \hline 
        \rowhighlight
         & \textbf{8B} & \textbf{4B} & \multicolumn{4}{c|}{\textbf{Qwen3-VL-8B-Instruct $\rightarrow$ Qwen3-VL-4B-Instruct}} & & \\
        \hline
        MMMU-Pro-4 &  62.03  & 48.93  & 55.79  & 56.43  & \textbf{56.86}  & \textbf{56.86} & \textbf{\color{darkgreen}+1.07}  & \textbf{\color{darkgreen}+0.43} \\
        LogicVista  & 60.01   & 47.10  & 55.80  & 56.43 & \textbf{58.70}  & \textbf{56.98} & \textbf{\color{darkgreen}+2.90} & \textbf{\color{darkgreen}+0.55} \\
        MathVerse-VD & 79.63  & 57.18  & 71.53 & 73.60 & \textbf{74.31 } & \textbf{73.89} & \textbf{\color{darkgreen}+2.78}  & \textbf{\color{darkgreen}+0.29} \\
        MathVerse-VO &  73.85  &  53.67  & 70.18 & 68.33 & \textbf{71.10}  & \textbf{68.53} & \textbf{\color{darkgreen}+0.92}  & \textbf{\color{darkgreen}+0.20}\\
        VisualPuzzles &  43.15  &  26.54  & \textbf{40.75} & 39.17 &  40.31 & \textbf{39.22} & \textbf{\color{darkred}-0.44} & \textbf{\color{darkgreen}+0.05} \\
        MathVision &  44.14  &  31.40  & 37.96 & 39.26 & \textbf{40.59} & \textbf{40.16} & \textbf{\color{darkgreen}+2.63} & \textbf{\color{darkgreen}+0.90} \\
        VlmsAreBlind & 66.79 &  61.94  & 64.46 & 64.82 &  \textbf{65.49} & \textbf{65.92} & \textbf{\color{darkgreen}+1.03} & \textbf{\color{darkgreen}+1.10} \\
        \hline
        \textbf{Average} &  61.37  &  46.68  & 56.64 & 56.86 & \textbf{58.12} & 5\textbf{7.27} & \textbf{\color{darkgreen}+1.56} & \textbf{\color{darkgreen}+0.50} \\
    \bottomrule
    \end{tabular}
     }
     \vspace{-10pt}
\end{table*}
\paragraph{Mitigating Destructive Gradient Interference.} While VGS effectively steers generation toward visual grounding, our analysis (Figure~\ref{fig: grad_angle_analysis}-(a)) reveals that tokens with the highest visual dependency exhibit \textbf{obtuse angles} ($> 90^\circ$) between the visual and language gradients. In these regimes, optimizing solely for visual grounding yields a negative projection onto the language gradient, effectively ``unlearning'' the language prior. Figure~\ref{fig: training loss curve} confirms this empirically: the Language Prior Matching loss ($\mathcal{L}_{\text{Lang}}$) for high-VDS tokens diverges significantly during training with $\mathcal{L}_{\text{VGS}}$ alone.

To prevent this, we introduce a \textbf{Language Preservation (LP)} regularizer. We apply this term selectively to the \textbf{top 30\% of tokens ranked by VDS} (Eq.~\ref{eq: VDS}; corresponding to Bins 7--9), where this gradient conflict is most prevalent. We define the trajectory-wise preservation loss as:
\begin{equation}
\label{eq:lp_term}
\small
\begin{split}
\ell_{\text{LP}}(\tau) \triangleq \frac{1}{|\tau|} \sum_{t=1}^{|\tau|} \Big( & \mathds{1}[\text{VDS}_t > Q_{0.7}] \\
& \hspace{-2em} \cdot D_{KL}\big( p_\text{S}^\theta(\cdot \mid \tau_{<t}, x) \parallel q_\text{T}(\cdot \mid \tau_{<t}, x) \big) \Big),
\end{split}
\end{equation}
where $Q_{0.7}$ denotes the \textbf{70th percentile threshold} of the Visual Dependency Score distribution. This term penalizes deviation from the teacher's language prior only when the visual steering signal is likely to be destructive. We incorporate this into our final trajectory-wise objective:
\begin{equation}
\label{eq:final_objective_pointwise}
\begin{split}
\ell_{\text{VGS-LP}}(\tau) &= \ell_{\text{VGS}}(\tau) + \lambda \ell_{\text{LP}}(\tau) \\
&= \ell_{\text{Standard}}(\tau) + \gamma \ell_{\text{Vis}}(\tau) + \lambda \ell_{\text{LP}}(\tau),
\end{split}
\end{equation}

where $\gamma$ acts as a primary steering coefficient (typically $\gamma \ge 1$) to drive visual adaptation, while $\lambda$ serves as a conservative regularization weight (e.g., $\lambda \approx 0.01$) to selectively prevent catastrophic forgetting.
The total loss is minimized over the expectation of this objective, scaled by the adaptive normalization factor $\eta_{\text{VGS}}(\gamma)$ in Eq. \ref{eq:beta_scaling}:
\begin{equation}
\label{eq:vgs_final_lp}
\mathcal{L}_{\text{VGS-LP}} = \eta_{\text{VGS}}(\gamma) \cdot \mathop{\mathbb{E}}\limits_{\tau \sim p_S^\theta(\cdot \mid I, x)} \left[ \ell_{\text{VGS-LP}}(\tau) \right].
\end{equation}
As demonstrated in Figure~\ref{fig: training loss curve}, this formulation successfully mitigates the divergence of the language prior, ensuring that the model retains its reasoning capabilities even in high-conflict visual regimes.

\section{Experimental Setup} \label{sec:experimental_setup}

\subsection{Models and Datasets} \label{subsec:models_datasets}

We conduct our training on the \href{https://huggingface.co/datasets/LMMs-Lab-Turtle/Vision-SR1-47K}{\textbf{Vision-SR1-47K}} dataset \cite{sr1}. This dataset consists of 47k triplets $(I, x, y)$, containing an image $I$, a question $x$, and a verifiable final answer $y$. For a controlled experimental setting, we avoid using off-the-shelf reasoning models where the training mixture is unknown. Instead, we train a reasoning teacher by fine-tuning the \href{https://huggingface.co/Qwen/Qwen3-VL-8B-Instruct}{\textbf{Qwen3-VL-8B-Instruct}} \cite{bai2025qwen3vltechnicalreport} using Group Relative Policy Optimization (GRPO)~\cite{grpo} for 2 epochs on Vision-SR1-47K dataset. Full implementation details are in Appendix \ref{app:teacher_imp}.
For the student policies, we utilize the smaller \href{https://huggingface.co/Qwen/Qwen3-VL-2B-Instruct}{\textbf{Qwen3-VL-2B-Instruct}} and \href{https://huggingface.co/Qwen/Qwen3-VL-4B-Instruct}{\textbf{Qwen3-VL-4B-Instruct}} \cite{bai2025qwen3vltechnicalreport}.  

We evaluate on seven vision-language reasoning benchmarks. For mathematical reasoning, we utilize \href{https://huggingface.co/datasets/MathLLMs/MathVision}{\textbf{MathVision}}~\cite{mathvision} and \href{https://huggingface.co/datasets/AI4Math/MathVerse}{\textbf{MathVerse}}~\cite{mathverse}, reporting results on the Visual Dominant (VD) and Visual Only (VO) subsets. For visual logic, we employ \href{https://github.com/Yijia-Xiao/LogicVista}{\textbf{LogicVista}}~\cite{logicvista}, \href{https://huggingface.co/datasets/neulab/VisualPuzzles}{\textbf{VisualPuzzles}}~\cite{visualpuzzles}, and \href{https://huggingface.co/datasets/XAI/vlmsareblind}{\textbf{VlmsAreBlind}}~\cite{vlmsareblind}. Finally, we test multidisciplinary reasoning using \href{https://huggingface.co/datasets/MMMU/MMMU_Pro}{\textbf{MMMU-Pro}}~\cite{yue2024mmmu} in its 4-option setting.

\subsection{Implementation details}
\paragraph{Prompt Format.} To ensure structural alignment, both the teacher (during GRPO) and the student (during distillation) utilize a unified system prompt. This enforces a strict reasoning format, requiring the model to explicitly delimit the internal reasoning trace from the final answer. The full prompt template is provided in Appendix~\ref{app:templates}.

\paragraph{Hyperparameters.} We optimize the objective in Eq.~\ref{eq:final_objective_pointwise} with a steering coefficient of $\gamma = 2.0$ and $\lambda = 0.01$ across all experiments. For the gradient normalization, we set $\eta(\gamma)=0.41$ for Qwen3-VL-2B-Instruct and $\eta(\gamma)=0.36$ for Qwen3-VL-4B-Instruct as justified in Appendix \ref{app:gradient_norm_eta}. All other training configurations were kept identical to the baseline for fair comparison. Full hyperparmater details and sensitivity analysis is provided in Appendix \ref{app:student_imp}, \ref{subsec:sensitivity}.

\section{Experimental Results}
\label{sec:exp}

We evaluate Visual Gradient Steering (VGS) by distilling a GRPO-trained Qwen3-VL-8B-Instruct teacher into 2B and 4B student models. As detailed in Table~\ref{tab: main}, VGS consistently outperforms the standard monolithic baseline across all seven vision-language benchmarks. The improvement is particularly pronounced when the teacher-student capacity gap is large. In the \textbf{Qwen3-VL-8B $\rightarrow$ Qwen3-VL-2B} setting, VGS achieves an average greedy accuracy (Acc@1) of \textbf{46.10\%}, surpassing the baseline (43.74\%) by a significant margin of \textbf{+2.37\%}. This gain is driven by substantial improvements in visually intensive tasks like VisualPuzzles (+3.68\%) and LogicVista (+3.35\%). This trend extends to the larger \textbf{Qwen3-VL-4B} student, where VGS improves average Acc@1 to \textbf{58.12\%} (+1.56\% over baseline), effectively closing the gap to the 8B teacher (61.37\%). 

Furthermore, we assess policy robustness via stochastic sampling (Acc@16), calculated as the average accuracy over 16 independent generations (temperature $T=1.0$), where VGS consistently yields higher average accuracy across both scales. \textit{We provide a computational overhead comparison in Appendix~\ref{app: compute_overhead} and the training dynamics in Section~\ref{sec:accuracy_curve}.
}
\section{Validating the Asymmetric Maturity Hypothesis}
A core premise of our approach is the \textbf{Asymmetric Maturity Hypothesis} (Sec.~\ref{sec:geometric_analysis}): we posit that visual grounding is the primary optimization bottleneck, while the language prior is already comparatively robust. To validate this, we perform an inverse steering experiment: \textit{What happens if we steer the gradient towards the Language Prior instead?}

We define the \textbf{Language Steering} objective symmetrically to VGS-LP. This objective steers the generation toward the unimodal language prior ($\mathcal{L}_{\text{Lang}}$) while strictly preserving visual grounding on high-dependency tokens:
\begin{align}
\label{eq:lang_steer_final}
\small
\mathcal{L}_{\text{Lang-Steer}} = & \eta_{\text{Lang}}(\gamma_{\text{Lang}}) \cdot \mathop{\mathbb{E}}\limits_{\tau \sim p_S^\theta(\cdot \mid I, x)} \Big[ \ell_{\text{Standard}}(\tau) \nonumber \\
& + \gamma_{\text{Lang}} \ell_{\text{Lang}}(\tau) + \lambda \ell_{\text{VP}}(\tau) \Big],
\end{align}
where $\eta_{\text{Lang}}(\gamma_{\text{Lang}})$ is the adaptive scaling factor, defined analogously to Eq.~\ref{eq:beta_scaling} but using the language gradient direction to maintain optimization stability:
\begin{equation}
\label{eq:eta_lang}
\small
\eta_{\text{Lang}}(\gamma) \triangleq \frac{\| \nabla_\theta \mathcal{L}_{\text{Standard}} \|_2}{\| \nabla_\theta \mathcal{L}_{\text{Standard}} + \gamma_{\text{Lang}} \nabla_\theta \mathcal{L}_{\text{Lang}} \|_2}.
\end{equation}
The term $\ell_{\text{VP}}(\tau)$ is the \textbf{Visual Preservation} regularizer, which penalizes deviations from the teacher's visual distribution $q^*_T$ (Eq.~\ref{eq:target_q}) specifically on tokens where visual grounding is most critical:
\begin{align}
\label{eq:vp_term}
\small
\ell_{\text{VP}}(\tau) & \triangleq \frac{1}{|\tau|} \sum_{t=1}^{|\tau|} \Big( \mathds{1}[\text{VDS}_t > Q_{0.7}] \nonumber \\
& \cdot D_{KL}\big( p_\text{S}^\theta(\cdot \mid \tau_{<t}, x) \parallel q^*_\text{T}(\cdot \mid \tau_{<t}, x) \big) \Big).
\end{align}
Similar to our VGS-LP formulation (Eq.~\ref{eq:final_objective_pointwise}), we set $\lambda$ to a small regularization value (i.e., 0.01). This ensures a symmetric comparison where both objectives steer a primary modality while safeguarding the other against catastrophic forgetting.

\begin{figure}[t]
	\centering
	\includegraphics[width=1.0\columnwidth]{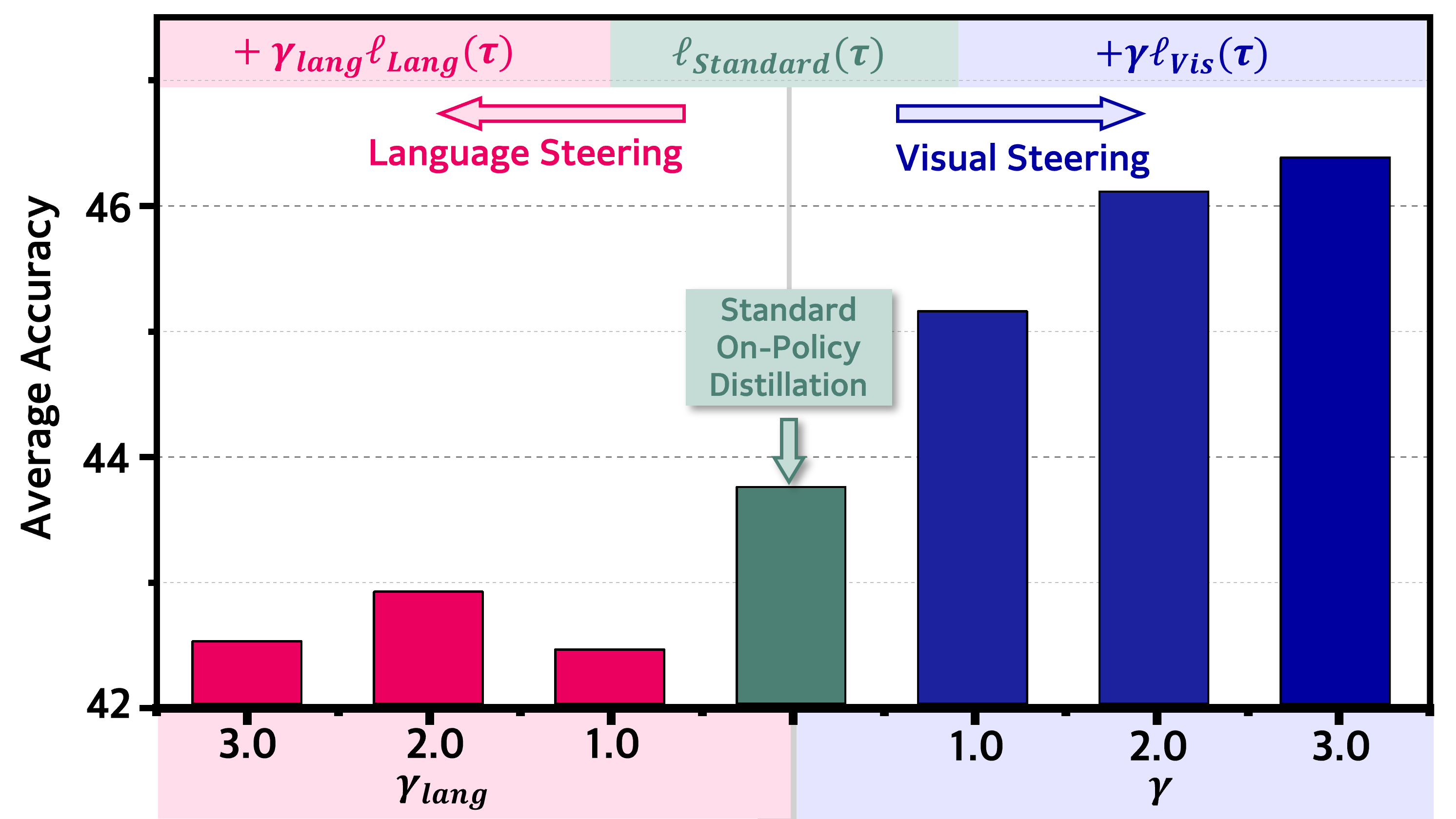}
\caption{\textbf{Validating the Asymmetric Maturity Hypothesis.} While steering toward the language matching reduces accuracy compared to the standard baseline, steering toward the visual subspace yields consistent performance gains.} 
    \label{fig: validate_hypthosis.}
\end{figure}

Figure~\ref{fig: validate_hypthosis.} compares the performance of Visual Steering ($\gamma > 0$) versus Language Steering ($\gamma_{\text{lang}} > 0$). As shown in the plot (where the y-axis represents the average accuracy across the 7 benchmarks listed in Table~\ref{tab: main}), increasing the visual steering coefficient $\gamma$ yields improvements, boosting the average from the baseline. In contrast, steering towards the language prior ($\gamma_{\text{lang}} > 0$) harms performance, with accuracy dropping below the baseline. This confirms that the model's language prior is already robust; forcing updates towards it consumes the optimization budget without gain. The visual modality, however, remains the critical bottleneck, responding positively to targeted gradient injection.

\begin{table*}[!t]
    \scriptsize
    \setlength{\extrarowheight}{3pt}
    \setlength{\tabcolsep}{6pt}
    \centering
\caption{\textbf{Ablation Results on RL + On-Policy Distillation.} We evaluate the effectiveness of integrating distillation objectives into the GRPO training loop using \textbf{Qwen3-VL-2B-Instruct}. Our Visual Gradient Steered (VGS) method consistently outperforms the standard monolithic distillation baseline when combined with RL.}
        \label{tab:grpo}
    \resizebox{1.0\textwidth}{!}{%
    \begin{tabular}{l|c|cc|cccc|cc}
    \toprule
        \multirow{2}{*}{\textbf{Benchmark}} &  \textbf{\shortstack{Initial \\ Student (2B)}}  & \multicolumn{2}{c|}{\textbf{GRPO}} & \multicolumn{2}{c}{\textbf{\shortstack{GRPO + Standard  \\ On-Policy Distillation}}} & \multicolumn{2}{c|}{\textbf{\shortstack{GRPO + VGS \\ On-Policy Distillation (ours)}}} & \multicolumn{2}{c}{\textbf{Improvement}} \\ [1ex]
        \cdashline{2-10}
                  &  \rule{0pt}{5ex} \shortstack{Acc@1 \\ (greedy)} & \shortstack{Acc@1 \\ (greedy)}  &  \shortstack{Acc@16 \\ (T=1.0)} &  \shortstack{Acc@1 \\ (greedy)}  & \shortstack{Acc@16 \\ (T=1.0)}  &  \shortstack{Acc@1 \\ (greedy)} & \shortstack{Acc@16 \\ (T=1.0)} & \shortstack{Acc@1 \\ (greedy)} & \shortstack{Acc@16 \\ (T=1.0)} \\
        \hline 
        MMMU-Pro-4 & 34.51  & 47.66  & 49.59  & 45.85  & 46.86  & \textbf{50.29}  & \textbf{49.26} & \textbf{\color{darkgreen}+4.44}  & \textbf{\color{darkgreen}+2.40} \\
        LogicVista  & 36.83  & 46.43   & 44.24  & 44.64  & 45.28  & \textbf{46.88}  & \textbf{46.01} & \textbf{\color{darkgreen}+2.24}  & \textbf{\color{darkgreen}+0.73} \\
        MathVerse-VD &  35.88  & 59.03   & 59.68  & 61.11  & 60.43  & \textbf{64.81}  & \textbf{61.44} & \textbf{\color{darkgreen}+3.70}  & \textbf{\color{darkgreen}+1.01} \\
        MathVerse-VO &  35.32  & 53.67   & 56.47  & \textbf{58.03}  & 55.89  & 57.80  & \textbf{57.70} & \textbf{\color{darkred}-0.23}  & \textbf{\color{darkgreen}+1.81} \\
        VisualPuzzles &  13.36  & 30.22   & 31.64  & 31.08 & 31.36  & \textbf{31.68}  & \textbf{32.23} & \textbf{\color{darkgreen}+0.60}  & \textbf{\color{darkgreen}+0.87}\\
        MathVision &  14.28  & 27.20   & 28.02  & 25.26  & 25.98  & \textbf{27.17}  & \textbf{27.52} & \textbf{\color{darkgreen}+1.91}  & \textbf{\color{darkgreen}+1.54} \\
        VlmsAreBlind & 49.03  & 49.61   & 50.16  & \textbf{51.90}  & 50.77  & 51.74  & \textbf{51.85} & \textbf{\color{darkred}-0.16}  & \textbf{\color{darkgreen}+1.08} \\
        \hline
        \textbf{Average} &  31.32  & 44.83  & 45.68  & 45.41  & 45.22 & \textbf{47.20}  & \textbf{46.57} & \textbf{\color{darkgreen}+1.79}  & \textbf{\color{darkgreen}+1.35}\\
    \bottomrule
    \end{tabular}
    }
\end{table*}

\begin{figure*}[t!]
	\centering
	\includegraphics[width=\textwidth]{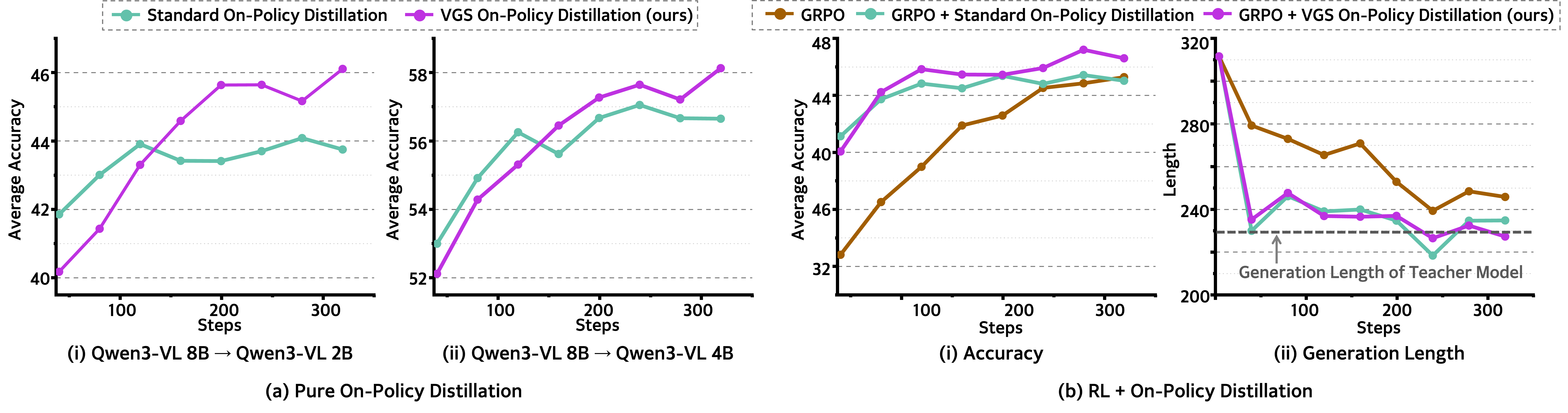}
\caption{\textbf{Training Dynamics Analysis of Pure Distillation and GRPO Settings.} \textbf{(a)} Average accuracy evolution during on-policy distillation for Qwen3-VL-2B and 4B students. VGS achieves higher accuracy throughout training compared to the standard monolithic baseline. \textbf{(b)} Analysis of the GRPO framework (using Qwen3-VL-2B). \textit{Left:} VGS combined with GRPO yields the highest accuracy trajectory, outperforming both pure GRPO and GRPO + Standard Distillation. \textit{Right:} While pure GRPO suffers from length explosion, both distillation methods regularize generation length to match the teacher's average; however, VGS achieves significantly higher accuracy within this efficient length constraint.}
    \label{fig: rl_dynamics}
\end{figure*}

\section{RL Fine-Tuning + On-Policy Distillation}
\label{sec:rl_finetuning}

As noted by \citet{gkd}, on-policy distillation can be naturally combined with Reinforcement Learning (RL) objectives, as both paradigms involve sampling trajectories directly from the student's current policy. We investigate whether our proposed steering towards the visual subspace ($\mathcal{L}_{\text{VGS-LP}}$, Eq. \ref{eq:final_objective_pointwise}) can enhance this synergy when fine-tuning with GRPO~\cite{grpo}.

Following the formulation of \citet{gkd}, we construct the unified training objective as a weighted convex combination of the RL reward and the distillation constraint. We define the total objective $\mathcal{J}_{\text{Total}}(\theta)$ as:
\begin{equation}
\label{eq:unified_objective}
\begin{split}
\mathcal{J}_{\text{Total}}(\theta) = \mathop{\mathbb{E}}\limits_{\{\tau_i\}_{i=1}^G \sim p_S^\theta} \Bigg[ \frac{1}{G} \sum_{i=1}^G \Big( & (1 - \alpha) \mathcal{O}_{\text{GRPO}}(\tau_i) \\
& - \alpha \ell_{\text{VGS-LP}}(\tau_i) \Big) \Bigg],
\end{split}
\end{equation}
where $\mathcal{O}_{\text{GRPO}}(\tau_i)$ is the standard GRPO surrogate objective (clipped advantage) without the KL penalty, and $\ell_{\text{VGS-LP}}$ serves as the visual distillation regularizer. We compare three training configurations on the Vision-SR1-47K dataset \cite{sr1}:
\begin{enumerate}
    \item \textbf{GRPO (Standard):} Pure RL optimization maximizing the correctness reward (equivalent to $\alpha=0$).
    \item \textbf{GRPO + Standard-KD:} RL regularized by the standard monolithic distillation loss. This corresponds to Eq.~\ref{eq:unified_objective} where $\ell_{\text{VGS-LP}}$ is replaced by $\ell_{\text{Standard}}$. We set $\alpha$ to 0.3
    \item \textbf{GRPO + VGS (Ours):} RL regularized by our visually steered objective in Eq.~\ref{eq:unified_objective}. We set $\alpha$ to 0.3.
\end{enumerate}

As shown in Table~\ref{tab:grpo}, while Standard-KD improves stability over pure GRPO, it suffers from the compromise effect, pulling the policy back towards the teacher's generic average. \textbf{GRPO + VGS} significantly outperforms both baselines. By using VGS as the regularizer, we enforce a constraint that is orthogonal to the RL reward: the RL objective optimizes for the \textit{correct answer}, while VGS ensures the \textit{reasoning process} remains visually grounded. This suggests that VGS is a versatile objective that can be plugged into various post-training paradigms to enforce perceptual fidelity. 

\section{Training Dynamics}
\label{sec:accuracy_curve}

In Figure~\ref{fig: rl_dynamics}, we present the average accuracy evolution across all 7 benchmark datasets during training. First, Figure~\ref{fig: rl_dynamics}-(a) compares the 2B and 4B students in the on-policy distillation setting. At both scales, our VGS approach consistently achieves a higher accuracy curve compared to the standard monolithic baseline, demonstrating the effectiveness of visual gradient steering.

Furthermore, Figure~\ref{fig: rl_dynamics}-(b) (left) displays the average accuracy in the RL (GRPO) + on-policy distillation setup that was introduced in Section \ref{sec:rl_finetuning}. While adding distillation accelerates training in general, our VGS yields a distinctively higher accuracy curve. Figure~\ref{fig: rl_dynamics}-(b) (right) illustrates the generation length dynamics. We observe that both distillation objectives constrain the student to converge toward the teacher's average generation length, preventing the length explosion seen in standard GRPO. Notably, VGS achieves significantly higher accuracy at this constrained length compared to Standard On-Policy Distillation, further validating the efficacy of visual gradient steering for reasoning efficiency.

\section{Related Works}
\label{sec:related_works}
\paragraph{Reinforcement Learning for Reasoning}
Expanding beyond foundational RLHF and preference optimization paradigms \cite{christiano2017deep, bai2022training, ouyang2022training, yoon2024tlcr, rafailov2023direct, yoon2025confpo}, recent large-scale reinforcement learning (RL) training across diverse tasks has significantly elevated the general reasoning capabilities of LLMs \cite{guo2025deepseek, team2025kimi, lambert2024tulu}. These advancements are predominantly driven by Reinforcement Learning from Verifiable Rewards (RLVR) frameworks \citep{qwen2_5_math, guo2025deepseek, grpo, yoon2025pacr, yoon2026pdcr}. 

However, a persistent challenge in standard RLVR environments is that outcome rewards are typically sparse, generated only at the very end of a token sequence. To resolve this credit assignment dilemma, several recent studies have developed process reward models (PRMs) \cite{li2025process, cheng2026stop, zhang2025progrm, lee2025rethinking} to evaluate step-by-step progressions within a generation; however, these PRMs typically demand specialized data curation and independent training pipelines. To circumvent reward sparsity without the overhead of an explicit PRM, on-policy distillation has emerged as a compelling alternative, leveraging dense, token-level supervision from a powerful teacher model \cite{yang2025qwen3, guha2025openthoughts, gkd}.

\paragraph{On-Policy Knowledge Distillation}
Knowledge Distillation (KD) has been widely adopted as a popular paradigm for developing efficient Small Large Language Models (SLMs) by leveraging supervision from larger teacher models \cite{yang2025qwen3, team2024gemma, guo2025deepseek, muralidharan2024compact, gu2024minillm, gkd}. \citet{gu2024minillm} demonstrated that reverse KL divergence is more effective than forward KL in preventing the student model from overestimating low-probability regions inherent in the teacher distribution. Furthermore, \citet{gkd} experimentally analyzed the trade-offs between on-policy KD, where the student model generates the training data, and off-policy settings, as well as the balance between reverse and forward KL objectives. In this work, we employ on-policy distillation with reverse KL divergence to enhance reasoning capabilities in instruction-tuned models with minimal distribution shift from initial student \cite{gu2024minillm}.

\paragraph{Vision-Focused Knowledge Distillation in MLLMs}
Knowledge distillation in the vision-language domain has recently garnered significant research attention as a key strategy for developing efficient, small-scale Multimodal Large Language Models (MLLMs). \citet{cai2025llavakd} utilized this approach to distill broad visual instruction-following capabilities from large-scale teachers into efficient student models. Furthermore, to address finer-grained perceptual alignment, \citet{jain2025elevating} introduced visual embedding distillation to transfer spatial knowledge, while \citet{kim2025compodistill} proposed an explicit attention alignment framework to enable student models to inherit the teacher’s compositional reasoning patterns. In contrast to computationally intensive methods relying on auxiliary encoders or explicit feature alignment, our approach minimizes complexity by steering the student's visual grounding directly through the teacher's decomposed output logits.

\section{Conclusion}
We identify a fundamental limitation in standard monolithic objectives: due to the \textit{asymmetric maturity} of MLLM modalities, the gradient acts as a passive bisector that fails to correct the weaker visual signal. We propose \textbf{Visual Gradient Steering (VGS)} to resolve this by explicitly prioritizing the visual subspace during optimization. Our experiments demonstrate that VGS not only outperforms standard distillation but also serves as a critical regularizer for RL (GRPO) fine-tuning. These findings confirm that breaking optimization symmetry, rather than treating modalities equally, is essential for unlocking faithful multimodal grounding. 

\section*{Acknowledgements}
This work was supported by Institute of Information \& communications Technology Planning \& Evaluation (IITP) grant funded by the Korea government(MSIT) (No. RS-2022-II0951, Development of Uncertainty-Aware Agents Learning by Asking Questions), and MSIT(Ministry of Science, ICT), Korea, under the Global Research Support Program in the Digital Field program(RS-2024-00436680) supervised by the IITP(Institute for Information \& Communications Technology Planning \& Evaluation). This project is supported by Microsoft Research Asia.



\section*{Impact Statement}
This paper presents work whose goal is to advance the field of Machine
Learning. There are many potential societal consequences of our work, none
which we feel must be specifically highlighted here.



\nocite{langley00}

\bibliography{example_paper}
\bibliographystyle{icml2026}

\newpage
\appendix
\onecolumn

\section{Analysis of the Adaptive Scaling Factor $\eta_{\text{VGS}}(\gamma)$}
\label{app:gradient_norm_eta}

In this section, we provide the analytical derivation and empirical justification for approximating the adaptive scaling factor $\eta_{\text{VGS}}(\gamma)$ as a constant.

\paragraph{Analytical Derivation.}
Recall the definition of the normalization factor from Eq.~\ref{eq:beta_scaling}. By expanding the squared norm in the denominator, we can express $\eta_{\text{VGS}}(\gamma)$ as a function of the gradient magnitudes and the angle $\phi$ between them:
\begin{equation}
\label{eq:eta_expansion}
\begin{aligned}
\eta_{\text{VGS}}(\gamma) & \triangleq \frac{\| \nabla \mathcal{L}_{\text{Standard}} \|_2}{\| \nabla \mathcal{L}_{\text{Standard}} + \gamma \nabla \mathcal{L}_{\text{Vis}} \|_2} \\
&= \frac{\| \nabla \mathcal{L}_{\text{Standard}} \|_2}{\sqrt{ A + B + C }},
\end{aligned}
\end{equation}
where:
\begin{align*}
A &= \| \nabla \mathcal{L}_{\text{Standard}} \|_2^2, \\
B &= \gamma^2 \| \nabla \mathcal{L}_{\text{Vis}} \|_2^2, \\
C &= 2\gamma \| \nabla \mathcal{L}_{\text{Standard}} \|_2 \| \nabla \mathcal{L}_{\text{Vis}} \|_2 \cos \phi,
\end{align*}
and $\phi$ represents the angle between the standard gradient $\nabla \mathcal{L}_{\text{Standard}}$ and the visual gradient $\nabla \mathcal{L}_{\text{Vis}}$.

\begin{figure}[h]
	\centering
	\includegraphics[width=0.5\columnwidth]{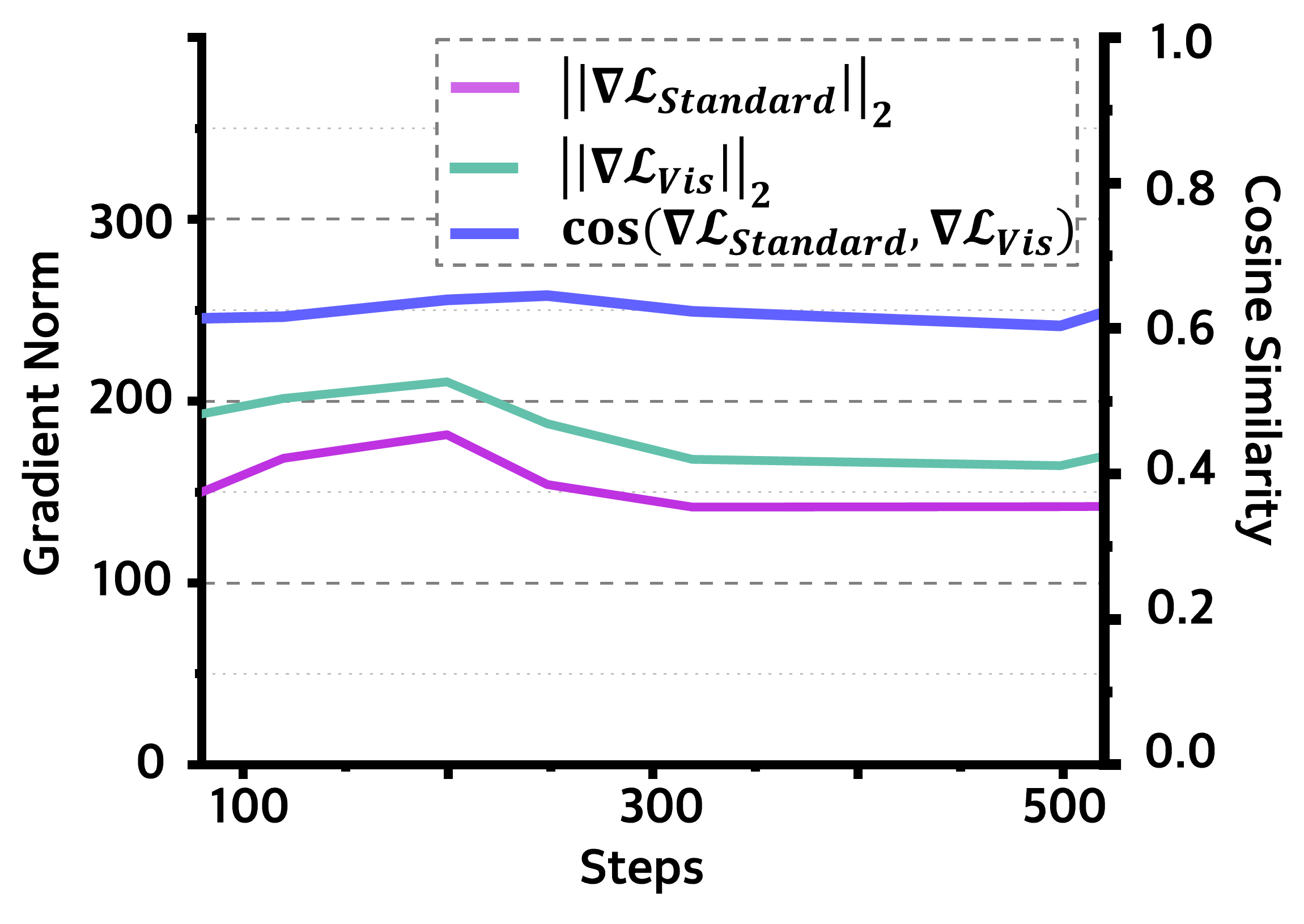}
\caption{\textbf{Evolution of Gradient Norm Components.} We track the magnitudes of the standard gradient (purple) and visual gradient (green), along with their cosine similarity (blue) for the \textbf{Qwen3-VL-2B-Instruct} student during \textbf{standard on-policy distillation}. All three components exhibit stability across training steps, justifying our approximation of the adaptive scaling factor $\eta_{\text{VGS}}(\gamma)$ as a fixed constant to reduce computational overhead.}
    \label{fig:grad_norm_analysis}
\end{figure}

Calculating Eq.~\ref{eq:eta_expansion} at every step incurs computational overhead due to the gradient norm computations. To mitigate this, we tracked the evolution of the constituent terms ($\| \nabla \mathcal{L}_{\text{Standard}} \|$, $\| \nabla \mathcal{L}_{\text{Vis}} \|$, and $\cos \phi$) throughout training. As illustrated in Figure~\ref{fig:grad_norm_analysis}, we observe that these values remain relatively constant across the training. Consequently, the resulting ratio $\eta_{\text{VGS}}(\gamma)$ exhibits high stability across the training.

Based on this observation, we treat $\eta_{\text{VGS}}(\gamma)$ as a fixed hyperparameter determined by the model architecture and $\gamma$. For our main experiments with $\gamma=2.0$, we utilize the pre-calculated average values:
\begin{itemize}
    \item \textbf{Qwen3-VL-2B-Instruct:} $\eta_{\text{VGS}} = 0.41$
    \item \textbf{Qwen3-VL-4B-Instruct:} $\eta_{\text{VGS}} = 0.36$
\end{itemize}
This approximation maintains the stability benefits of gradient normalization without the runtime cost.

\clearpage
\section{Implementation Details}
\label{app:implementation}

\subsection{Teacher Model Training}
\label{app:teacher_imp}
To generate high-quality reasoning trajectories for distillation, we train a teacher model using Group Relative Policy Optimization (GRPO). We initialize the teacher with Qwen3-VL-8B-Instruct and fine-tune it on the \texttt{Vision-SR1-47K} dataset for 2 epochs. This reinforcement learning stage ensures the teacher provides robust, verifiable reasoning chains rather than just final answers. We perform all GRPO experiments using the TRL codebase. The specific hyperparameters for the teacher training phase are detailed in Table~\ref{tab:teacher_hype}. 

\begin{table}[h!]
\small
\centering
\caption{\textbf{Hyperparameters for Teacher Model Training (GRPO).} The teacher model (Qwen3-VL-8B-Instruct) is trained for 2 epochs on Vision-SR1-47K.}
\begin{tabular}{ll}
\hline
\textbf{Hyperparameter} & \textbf{Value} \\
\hline
\rowcolor{gray!20} \multicolumn{2}{l}{\textit{Optimization}} \\
Optimizer           & AdamW \\
Learning Rate       & 1e-6 \\
Weight Decay        & 1e-2 \\
LR Schedule         & Constant \\
Epochs              & 2 \\
Global Batch Size   & 128 \\
Freeze Vision Encoder & False \\
\hline
\rowcolor{gray!20} \multicolumn{2}{l}{\textit{RL / Rollout Config}} \\
Rollout Batch Size  & 512 \\
Rollout Size ($G$)  & 8 \\
Rollout Temperature & 1.0 \\
Rollout Top-p       & 0.99 \\
Max Input Prompt Length & 12800 \\
Max Response Length & 2048 \\
Use KL Loss         & False \\
\hline
\end{tabular}
\label{tab:teacher_hype}
\end{table}

\subsection{Training Framework and Hyperparameters}
\label{app:student_imp}
We perform all distillation experiments using the TRL codebase. To ensure a controlled environment, we initialize all student models (Qwen3-VL-2B-Instruct and Qwen3-VL-4B-Instruct) from their standard instruction-tuned checkpoints and train them to match the GRPO-tuned Qwen3-VL-8B-Instruct teacher. To ensure a strictly fair comparison, we maintain a consistent configuration across all baselines (Standard Monolithic Distillation, Visual Gradient Steering) and ablation settings. Common hyperparameters for the optimizer, generation (rollout), and model constraints are detailed in Table~\ref{tab:hype}. All experiments were conducted on a single node equipped with 8 $\times$ NVIDIA A100 80GB GPUs.

\begin{table}[h!]
\small
\centering
\caption{\textbf{Key hyperparameters} used for On-Policy Distillation training and evaluation.}
\begin{tabular}{ll}
\hline
\textbf{Hyperparameter} & \textbf{Value} \\
\hline
\rowcolor{gray!20} \multicolumn{2}{l}{\textit{Optimization \& Training}} \\
Optimizer           & AdamW \\
Learning Rate       & 1e-6 \\
Weight Decay        & 1e-2 \\
LR Schedule         & Constant \\
Epochs              & 1 \\
Global Batch Size   & 512 \\
Freeze Vision Encoder & False \\
\hline
\rowcolor{gray!20} \multicolumn{2}{l}{\textit{On-Policy Rollout}} \\
Rollout Temperature & 1.0 \\
Rollout Top-p       & 1.0 \\
Max Input Prompt Length & 16384 \\
Max Response Length & 2048 \\
\hline
\rowcolor{gray!20} \multicolumn{2}{l}{\textit{VGS Specific (Ours)}} \\
Steering Coefficient ($\gamma$) & 2.0 \\
Regularization Weight ($\lambda$) & 0.01 \\
\hline
\end{tabular}
\label{tab:hype}
\end{table}

\subsection{Sensitivity Analysis for Visual Gradient Steering (VGS) Specific Hyperparameters}
\label{subsec:sensitivity}

\begin{figure}[h]
	\centering
	\includegraphics[width=0.8\columnwidth]{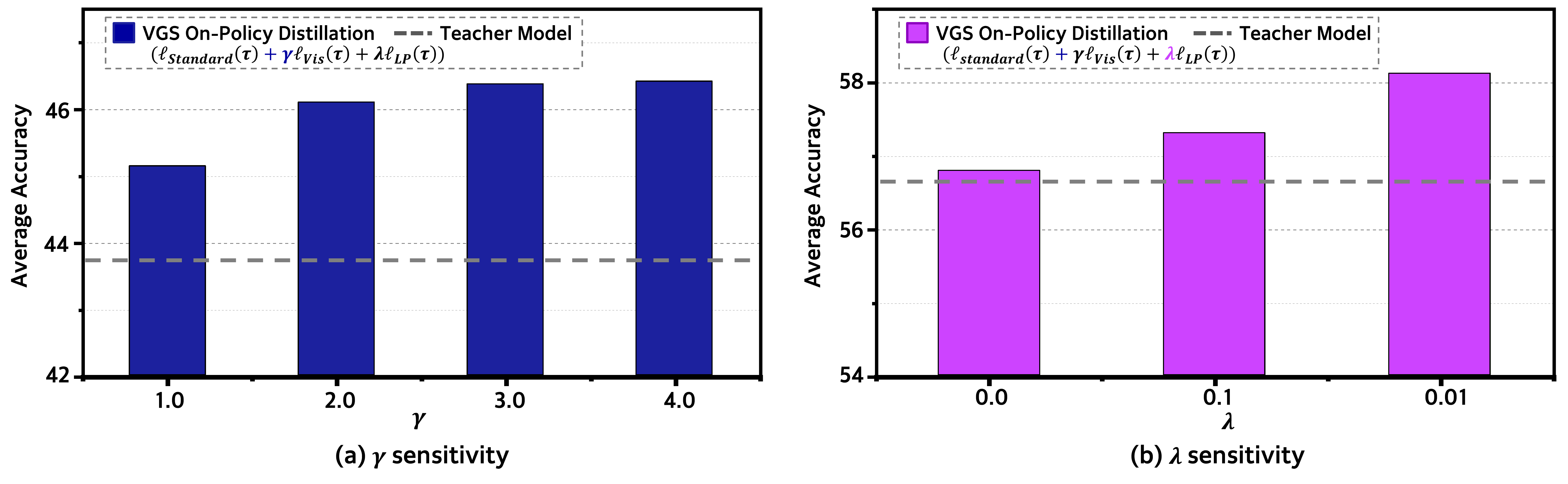}
\caption{\textbf{Hyperparameter Sensitivity Analysis.} We evaluate the impact of key hyperparameters on the average accuracy of the student model. (a) Varying the steering coefficient $\gamma$ (on Qwen3-VL-2B-Instruct): The method shows robustness across a wide range of $\gamma$ values, consistently outperforming the Standard On-Policy Distillation baseline (dashed grey line). (b) Varying the language preservation weight $\lambda$ (on Qwen3-VL-4B-Instruct): A small non-zero regularization (e.g., $\lambda=0.01$) yields the highest performance, confirming that the Language Preservation (LP) term effectively mitigates destructive interference without suppressing the visual signal.}
\label{fig:sensitivity}
\end{figure}

We analyze the sensitivity of $\mathcal{L}_{\text{VGS-LP}}$ (Eq. \ref{eq:final_objective_pointwise})  to the steering coefficient $\gamma$ and the language preservation weight $\lambda$ (Figure~\ref{fig:sensitivity}). First, regarding $\gamma$ (with $\lambda$ fixed at 0.01), VGS consistently outperforms standard on-policy distillation across a broad range of values, demonstrating that the benefits of visual steering are robust to hyperparameter variations. Second, regarding $\lambda$ ($\gamma$ fixed at 2.0), results confirm that omitting the regularizer ($\lambda=0$) is ineffective, as noted in Section~\ref{sec:method}, due to destructive gradient interference that causes the unlearning of the language prior in high-dependency regimes. However, assigning a small non-zero magnitude (e.g., $\lambda \in \{0.01, 0.1\}$) is sufficient to effectively preserve the language prior and yield performance consistently superior to the standard baseline.

\subsection{Prompt Template for Training and Inference}
\label{app:templates}

To ensure structural alignment between the teacher and student policies, we utilize a unified system prompt across all training stages (Teacher GRPO, Student On-Policy Distillation, and VGS). This prompt enforces a strict Chain-of-Thought (CoT) format, requiring the model to explicitly delimit its reasoning process within \texttt{<reason>} and \texttt{</reason>} tags before generating the final answer. The standard template used is as follows:

\definecolor{rliableblue}{HTML}{77AADD}
\newcommand{\placeholder}[1]{\textcolor{red}{\{#1\}}}

\tcbset{
  reasoningbox/.style={
    enhanced,
    colback=white,
    colframe=rliableblue!80!black,
    colbacktitle=rliableblue!80!black,
    coltitle=white,
    fonttitle=\bfseries,
    boxrule=0.7pt,
    arc=2mm,
    left=3mm,
    right=3mm,
    top=2mm,
    bottom=2mm,
    boxsep=1.5pt
  }
}
\begin{tcolorbox}[reasoningbox,title=Reasoning Template for Training and Inference]
\textbf{SYSTEM:} \\
A conversation between user and assistant. The user asks a question, and the assistant solves it. The assistant first thinks about the reasoning process in the mind and then provides the user with the answer. The reasoning process should be enclosed within \textless reason\textgreater \textless/reason\textgreater tags. The final answer MUST BE put in \textbackslash boxed\{\}. \\[0.5em]

\textbf{USER:} \\
\placeholder{image} \\
\placeholder{question} \\[0.8em]
\end{tcolorbox}

\clearpage

\section{Computational Cost Overhead}
\label{app: compute_overhead}

\begin{figure}[t]
	\centering
    \includegraphics[width=0.3\columnwidth]{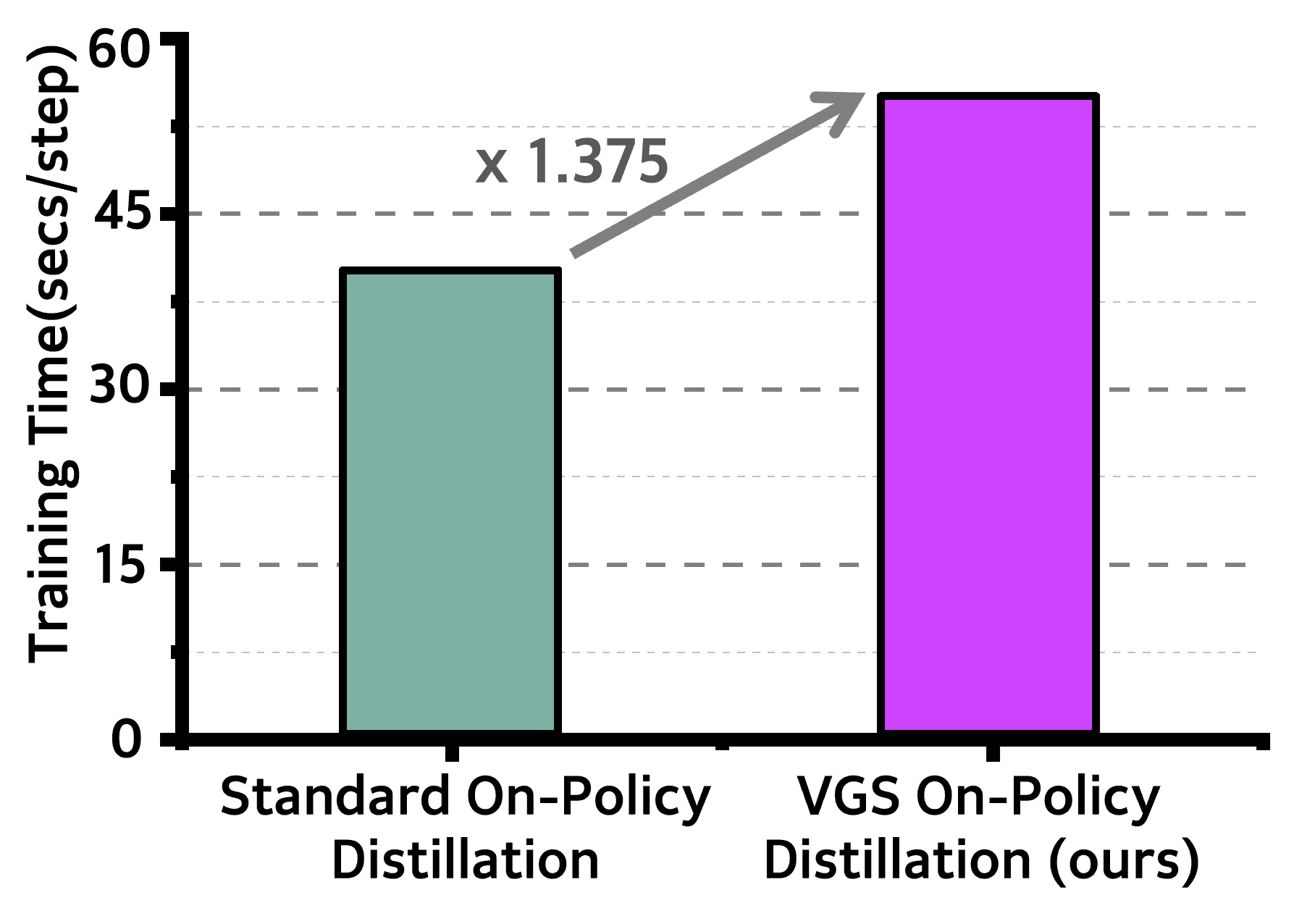}
    \caption{\textbf{Training Cost Comparison.} VGS incurs a modest $1.375\times$ increase in training time per step due to the dual forward passes required to isolate the visual gradient.}
    \label{fig:computation_overhead}
\end{figure}


We analyze the computational efficiency of our approach in Figure~\ref{fig:computation_overhead}. VGS introduces a moderate overhead, increasing the training time per step by approximately $1.375\times$ compared to standard on-policy distillation. This increase stems from the necessity of an additional non-image-conditioned forward pass to calculate the decomposed visual gradient. We consider this a minimal and acceptable cost given the significant performance improvements yielded by the steering mechanism.

\begin{figure*}[h!]
    \centering
    \includegraphics[width=\textwidth]{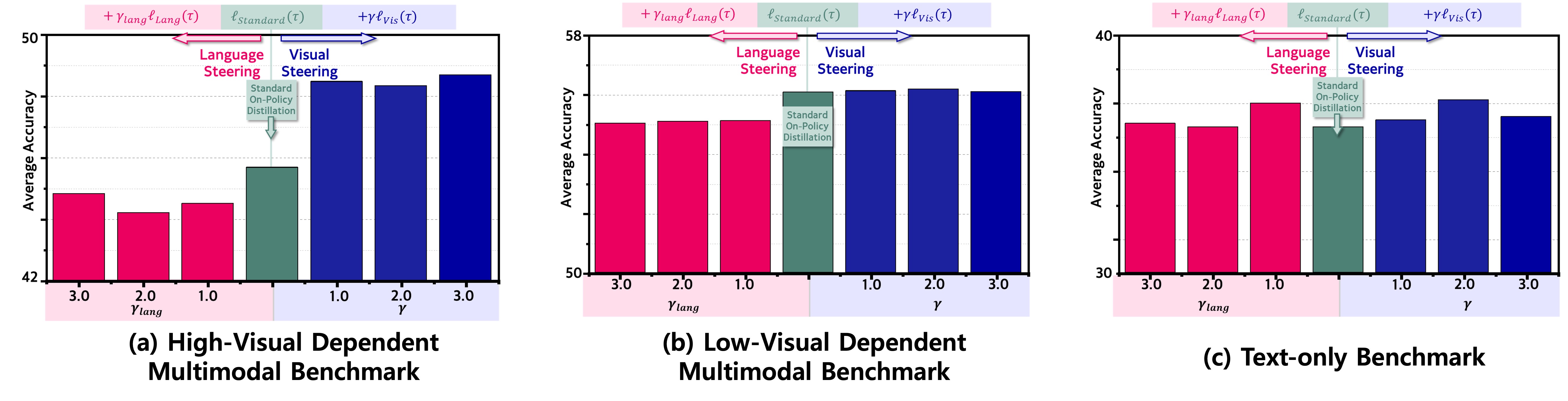}
    \caption{\textbf{Impact of optimization steering across varying levels of
    visual dependency.} Average accuracy comparison between Language Steering,
    Standard On-Policy Distillation, and Visual Gradient Steering. \textbf{(a)} On
    High-Visual Dependent benchmarks, prioritizing the visual subspace (VGS)
    significantly outperforms the baseline, whereas leaning on the language prior
    actively degrades performance. \textbf{(b)} On Low-Visual Dependent and
    \textbf{(c)} pure Text-Only benchmarks, performance remains uniform across all
    methods, confirming that VGS yields gains where visual grounding is the primary
    bottleneck without compromising the model's core textual reasoning capabilities.}
    \label{fig:visual_dependency}
\end{figure*}

\section{When is Visual Gradient Steering Most Effective?}
\label{app:effectiveness}

To analyze when Visual Gradient Steering (VGS) yields the largest benefit, we evaluate our models across benchmarks categorized by their reliance on visual context~\citep{wang2025perception}. Figure~\ref{fig:visual_dependency} compares Language Steering, Standard On-Policy Distillation, and VGS as a function of the steering coefficient, grouped by visual dependency level.

\begin{itemize}
    \item \textbf{High Vision-Dependency (MMMU-Pro, LogicVista).} A clear upward
    trend emerges (Language Steering $<$ Standard OPD $<$ VGS), where prioritizing the visual subspace yields consistent gains.

    \item \textbf{Low Vision-Dependency (Geo3K, We-Math).} The performance
    difference across methods is minimal. When textual prompts contain sufficient
    information to deduce the answer, the optimization trajectory has minimal impact.

    \item \textbf{Text-Only (MATH500, AIME25, OlympiadBench).} All methods
    yield uniform performance. This demonstrates that prioritizing the visual
    subspace during distillation does not necessary compromise the model's core text-only reasoning capability.
\end{itemize}

In summary, VGS consistently outperforms standard OPD on tasks demanding strong multimodal grounding, without compromising performance on low-dependency or text-only tasks.

\clearpage
\section{Adaptive Token-Level Visual Gradient Steering}
\label{app:adaptive_vgs}

Our main formulation applies a single, fixed steering coefficient $\gamma$ across all tokens. A natural extension is to let the visual emphasis vary at the token level, scaling the steering strength according to how much each token relies on visual evidence. To this end, we explore a fully adaptive Visual Gradient Steering (VGS) objective that dynamically scales the steering coefficient $\gamma_t$ based on each token's Visual Dependency Score ($\mathrm{VDS}_t$, Eq.~\ref{eq: VDS}).

Formally, modifying our objective in Eq.~\ref{eq:final_objective_pointwise}, the adaptive objective is defined as:
\begin{equation}
    \ell_{\text{Adaptive}}(\tau) \triangleq
    \frac{1}{|\tau|} \sum_{t=1}^{|\tau|}
    \Big( \ell_{\text{Standard}}(\tau) + \gamma_t\, \ell_{\text{Vis}}(\tau)
    + \lambda\, \ell_{\text{LP}}(\tau) \Big),
    \label{eq:adaptive_vgs}
\end{equation}
where the token-level steering coefficient $\gamma_t$ adjusts dynamically:
\begin{equation}
    \gamma_t =
    \begin{cases}
        0, & \mathrm{VDS}_t \leq Q_{0.4} \quad \text{(no visual correction)}, \\[2pt]
        \gamma/2, & Q_{0.4} < \mathrm{VDS}_t \leq Q_{0.7} \quad \text{(moderate correction)}, \\[2pt]
        \gamma, & \mathrm{VDS}_t > Q_{0.7} \quad \text{(maximum visual steering)},
    \end{cases}
    \label{eq:adaptive_gamma}
\end{equation}
with $Q_{0.4}$ and $Q_{0.7}$ denoting the 40th and 70th percentile thresholds of
the Visual Dependency Score distribution. Intuitively, this allocates the steering budget toward tokens that genuinely require visual grounding, while leaving low-dependency tokens unchanged.

As shown in Table~\ref{tab: token_level}, this adaptive formulation slightly
outperforms our original fixed-scale approach (i.e., $\gamma = 2.0$), achieving the highest overall average accuracy. This indicates that matching the steering
strength to each token's visual demand could provide a modest additional gain over applying a uniform coefficient.

\begin{table*}[!t]
    \scriptsize
    \setlength{\extrarowheight}{0.5pt}
    \setlength{\tabcolsep}{6pt}
    \centering
\caption{\textbf{Main Results on Vision-Language Reasoning Benchmarks.} We compare the distillation performance of Visual Gradient Steering (VGS) against the standard monolithic baseline. All student models (2B and 4B) are distilled from the same Qwen3-VL-8B-Instruct teacher trained with GRPO. VGS consistently outperforms the standard approach across all benchmarks, achieving higher accuracy in greedy decoding (Acc@1).}
        \label{tab: token_level}
    \resizebox{1.0\textwidth}{!}{%
    \begin{tabular}{l|cc|ccc}
    \toprule
        \multirow{2}{*}{\textbf{Benchmark}} & \textbf{\shortstack{Teacher \\ (GRPO trained)}} &  \textbf{\shortstack{Initial \\ Student}} & \textbf{\shortstack{Standard  \\ On-Policy Distillation}} & \textbf{\shortstack{VGS \\ On-Policy Distillation (ours)}} & \textbf{\shortstack{Adaptive-VGS \\ On-Policy Distillation (ours)}} \\[1ex]
        \cdashline{2-6}
                  &  \rule{0pt}{5ex}  \shortstack{Acc@1 \\ (greedy)} & \shortstack{Acc@1 \\ (greedy)}  &  \shortstack{Acc@1 \\ (greedy)} & \shortstack{Acc@1 \\ (greedy)} & \shortstack{Acc@1 \\ (greedy)} \\
        \hline 
        \rowhighlight
        & \textbf{8B} & \textbf{2B} & \multicolumn{3}{c}{\textbf{Qwen3-VL-8B-Instruct $\rightarrow$ Qwen3-VL-2B-Instruct}} \\
        \hline
        MMMU-Pro-4 & 62.03  & 34.51  & 45.83  & \underline{48.07}  & \textbf{48.40} \\
        LogicVista  &  60.01  & 36.83  & 45.53  & \textbf{48.88}  & \underline{48.31} \\
        MathVerse-VD & 79.63  & 35.88  & 56.02 & \underline{58.10}  & \textbf{59.72} \\
        MathVerse-VO &  73.85  & 35.32  & 54.59 & \underline{56.19}  & \textbf{57.11} \\
        VisualPuzzles & 43.15 & 13.36  & 28.08 &  \underline{30.64} & \textbf{33.99} \\
        MathVision &  44.14  & 14.28  & 24.14 &  \textbf{25.59} & \underline{25.16} \\
        VlmsAreBlind &  66.79  & 49.03  & 51.86 & \textbf{54.11} & \underline{53.00} \\
        \hline
        \textbf{Average} &  61.37  & 31.32  &  43.74 & \underline{46.10}  & \textbf{46.53} \\
    \bottomrule
    \end{tabular}
     }
\end{table*}

\clearpage
\section{Qualitative Comparisons of Generated Reasoning}
\label{sec:qualitative}

In this section, we present side-by-side comparisons between our proposed \textbf{Visual Gradient Steering (VGS)} and the standard monolithic on-policy distillation baseline. All examples presented were generated by \texttt{Qwen3-VL-2B-Instruct} distilled from the \texttt{Qwen3-VL-8B-Instruct} teacher. These examples highlight the core benefit of explicitly steering gradients toward the visual subspace. As observed in the figures below, the standard baseline often exhibits a specific failure mode where it generates logically coherent reasoning chains that are premised on incorrect visual extraction (highlighted in \textcolor{red}{\textbf{red}}). In contrast, VGS guides the student to accurately ground its reasoning in the visual input (highlighted in \textcolor{teal}{\textbf{green}}).

\section{Limitations}
\label{sec:limitations}

While Visual Gradient Steering (VGS) significantly improves reasoning reliability, we acknowledge two primary limitations.

\paragraph{Training Throughput Overhead.}
The core mechanism of VGS requires decomposing the teacher's output distribution, which necessitates an additional forward pass without visual inputs at each training step. As analyzed in Section~\ref{app: compute_overhead}, this incurs a computational overhead of approximately $1.375\times$ per step compared to standard distillation. 

\paragraph{Dependence on Teacher Calibration.}
VGS operates on the premise that the teacher's multimodal and unimodal distributions are distinguishable. It relies on the \textit{contrastive signal} between the two to steer the student. Consequently, if the teacher model suffers from mode collapse where the visual input exerts no influence on the output distribution (i.e., $\nabla \mathcal{L}_{\text{Vis}} \approx 0$), VGS effectively reduces to standard distillation. Our method enhances the \textit{transfer} of visual grounding but cannot rectify fundamental perceptual blindness present in the teacher itself.

\begin{figure}[!p]
\centering
\vfill
\begin{promptbox}[Generated Sample 1 by Standard On-Policy Distillation]{lightorange}{prompt:orange}
\includegraphics[width=0.35\textwidth]{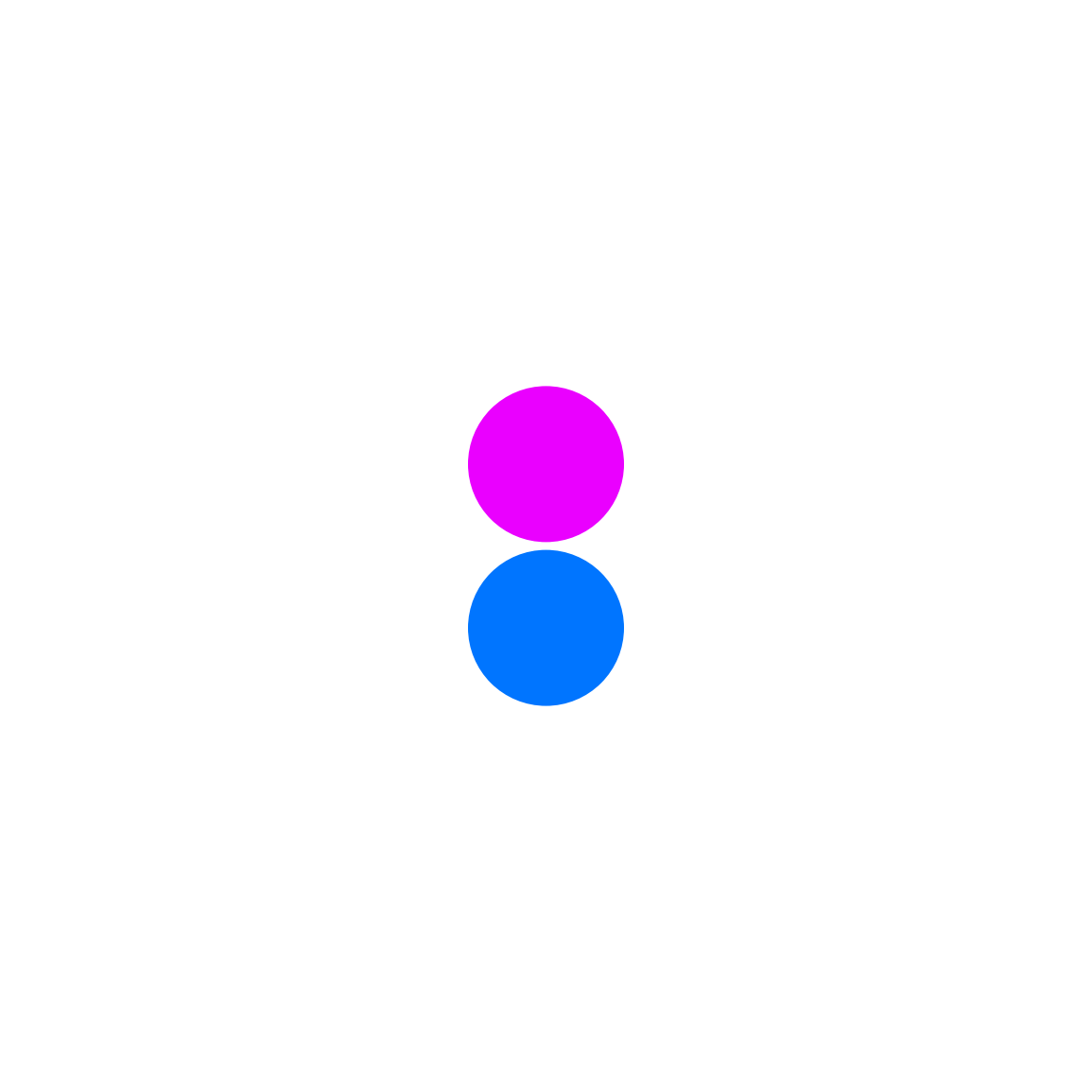}

\textbf{Question}: Are the two circles touching each other? Answer is Yes/No
\\
\colorbox{yellow!30}{\textbf{Ground truth: No}}

\textbf{Generation:}

\textbf{$<$reason$>$} \\
The two circles are positioned one above the other, \textcolor{red}{with no visible gap between them. Since they are touching at a single point (the bottom circle touches the top circle at their overlapping edge), they are considered touching.}
\\
\textbf{$<$/reason$>$} \boxed{\text{Yes}}

\end{promptbox}
\vfill
\end{figure}

\begin{figure}[!p]
\centering
\vfill
\begin{promptbox}[Generated Sample 1 by VGS On-Policy Distillation (ours)]{mintgreen}{prompt:orange}
\includegraphics[width=0.35\textwidth]{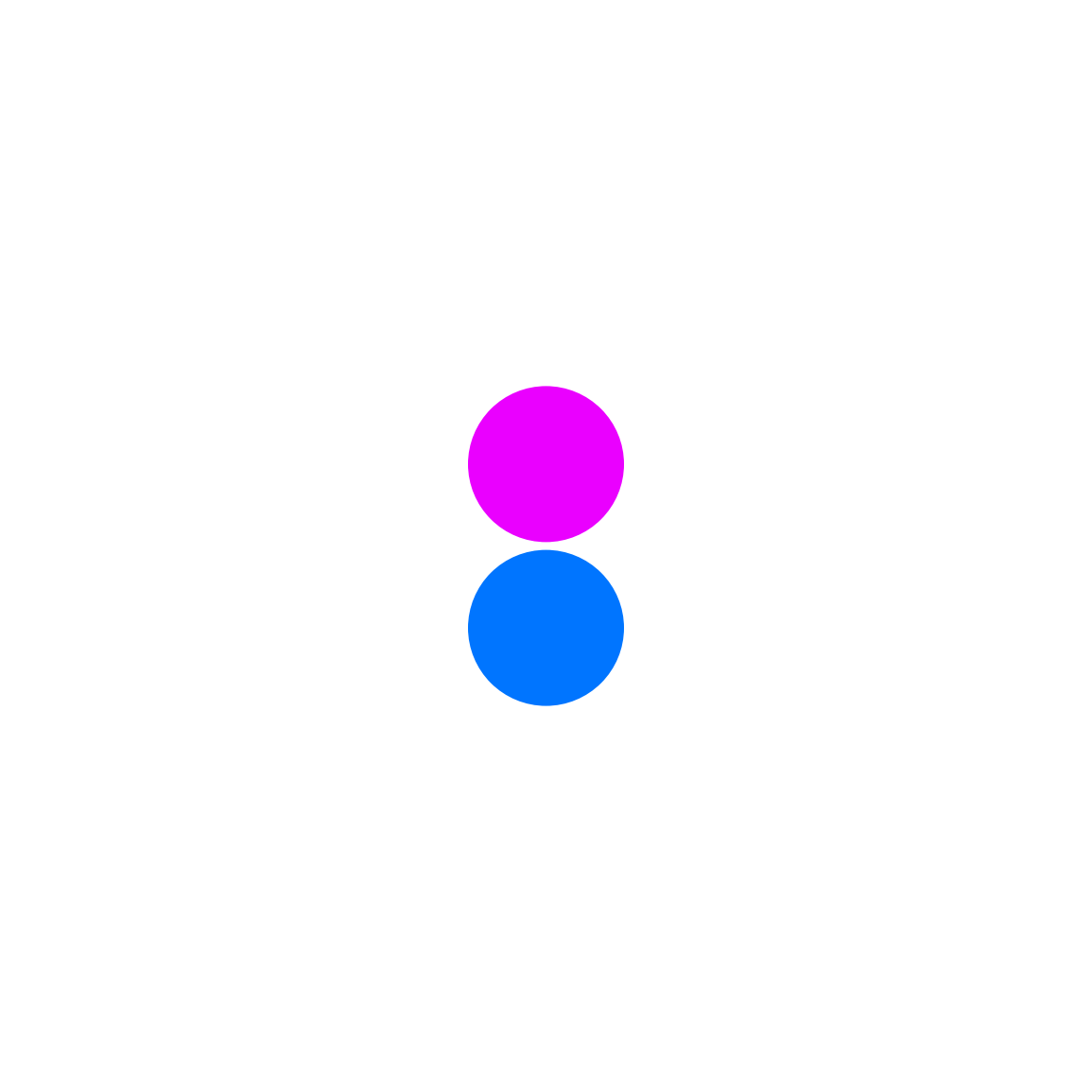}
\\
\textbf{Question}: Are the two circles touching each other? Answer is Yes/No
\\
\colorbox{yellow!30}{\textbf{Ground truth: No}}

\textbf{Generation:}

\textbf{$<$reason$>$} \\
The two circles in the image are positioned one above the other, \textcolor{teal}{but they do not touch each other. There is a clear space between them. Touching would mean that their edges meet at a point, but here they are separated by a gap.}
\\
\textbf{$<$/reason$>$}
\boxed{\text{No}}

\end{promptbox}
\vfill
\end{figure}

\begin{figure}[!p]
\centering
\vfill
\begin{promptbox}[Generated Sample 2 by Standard On-Policy Distillation]{lightorange}{prompt:orange}
\includegraphics[width=0.3\textwidth]{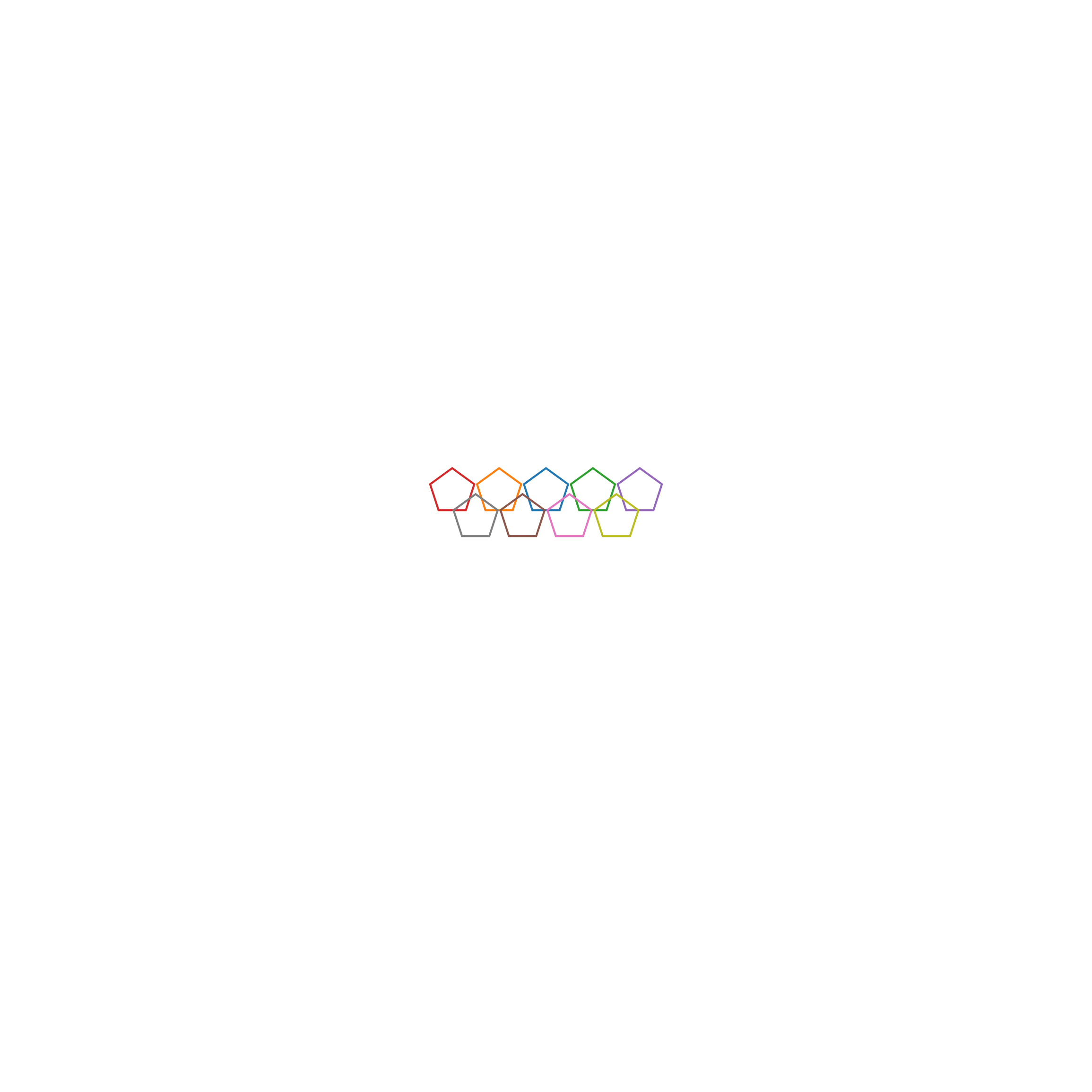}
\\
\textbf{Question}: Count the pentagons in the image. Answer with a number in curly brackets e.g., $\backslash$boxed\{3\}.
\\
\colorbox{yellow!30}{\textbf{Ground truth: 9}}

\textbf{Generation:}

\textbf{$<$reason$>$} \\
\textcolor{red}{The image shows a classic Olympic rings logo, which consists of five interlocking pentagons.} Each pentagon is a distinct color: red, orange, blue, green, and purple. There are no other pentagons in the image.
\\
Therefore, the total number of pentagons is 5.
\\
\textbf{$<$/reason$>$}
\boxed{\text{5}}

\end{promptbox}
\vfill
\end{figure}

\begin{figure}[!p]
\centering
\vfill
\begin{promptbox}[Generated Sample 2 by VGS On-Policy Distillation (ours)]{mintgreen}{prompt:orange}
\includegraphics[width=0.3\textwidth]{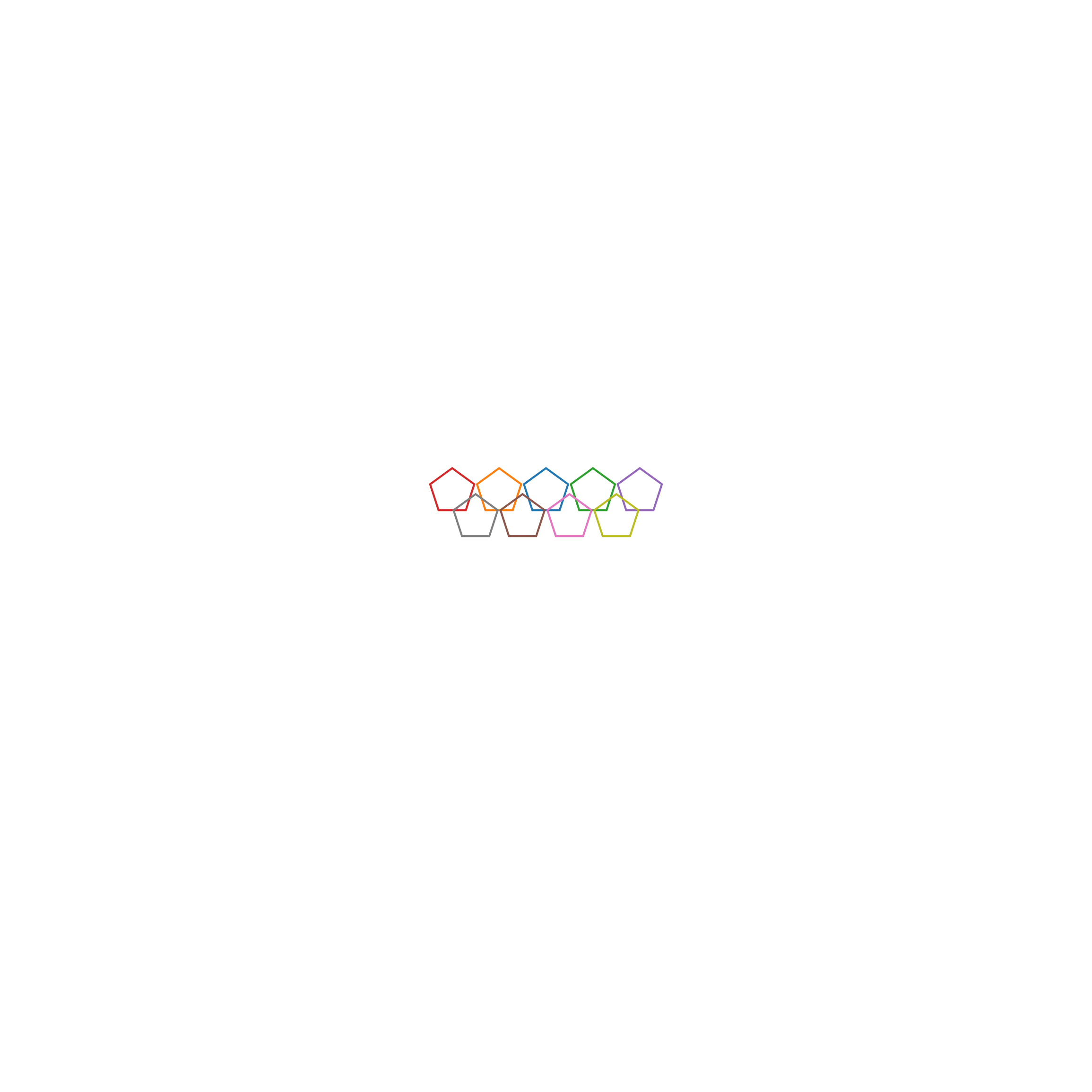}

\textbf{Question}: Count the pentagons in the image. Answer with a number in curly brackets e.g., $\backslash$boxed\{3\}.
\\
\colorbox{yellow!30}{\textbf{Ground truth: 9}}

\textbf{Generation:}

\textbf{$<$reason$>$} \\
Count the pentagons in the image. The image shows a row of pentagons, each colored differently. Count them one by one:
\\
\textcolor{teal}{
1. Red pentagon \\
2. Orange pentagon \\
3. Blue pentagon\\
4. Green pentagon\\
5. Purple pentagon\\
6. Gray pentagon\\
7. Brown pentagon\\
8. Pink pentagon\\
9. Yellow pentagon}
\\
There are 9 pentagons in total.
\\
\textbf{$<$/reason$>$} \boxed{\text{9}}

\end{promptbox}
\vfill
\end{figure}

\begin{figure}[!p]
\centering
\vfill
\begin{promptbox}[Generated Sample 3 by Standard On-Policy Distillation]{lightorange}{prompt:orange}
\includegraphics[width=0.3\textwidth]{qual_figures/MMMUPro4_image_00000456.png}
\\
\textbf{Question}: Which of the following statements is most solidly supported by the information contained in the map shown in the image?

\textbf{Choices}:\\
(a) Cossack communities devoutly adhered to a conservative strain of Russian Orthodox Christianity.\\
(b) Cossack hosts frequently rebelled against the centralizing authority of the Russian Empire. \\
(c) Cossack military units consist of highly skilled horse  back warriors. \\ 
(d) Cossack ethnic identity was shaped by intermarriage with non-Russian peoples on the empire's frontiers.
\\
\colorbox{yellow!30}{\textbf{(d)}}

\textbf{Generation:}

\textbf{$<$reason$>$} \\
The map shown is a \"Map of major Cossack hosts of the former Russian Empire.\" It highlights various regions where Cossack communities were centered. \textcolor{red}{However, the map does not provide any information about:}
\\ \\
- Religious practices (A) – The map does not indicate any religious affiliations or beliefs.
\\
- Rebellion against central authority (B) – The map does not show any evidence of rebellion or resistance.
\\
- Military skills (C) – The map does not provide any information about the military capabilities or training of Cossack units.
\\
- Ethnic identity (D) – The map does not indicate any information about intermarriage or ethnic mixing.
\\ \\
Since the map only shows the locations of Cossack hosts, and none of the options can be directly supported by the map's content, we must choose the option that is most likely to be true based on general historical knowledge of the Cossacks.
\\ \\
The Cossacks were a semi-nomadic ethnic group that played a significant role in the Russian Empire's military and political history. They were known for their martial prowess and were often involved in frontier conflicts. While the map does not provide direct evidence, option C (\"Cossack military units consisted of highly skilled horseback warriors\") is a well-documented fact about the Cossacks, and \textcolor{red}{it is the most solidly supported by general historical knowledge, even though it is not directly stated on the map.}
\\
Therefore, the best answer is C.
\textbf{$<$/reason$>$}
\boxed{\text{(c)}}

\end{promptbox}
\vfill
\end{figure}

\begin{figure}[!p]
\centering
\vfill
\begin{promptbox}[Generated Sample 3 by VGS On-Policy Distillation (ours)]{mintgreen}{prompt:orange}
\includegraphics[width=0.3\textwidth]{qual_figures/MMMUPro4_image_00000456.png}

\textbf{Question}: Which of the following statements is most solidly supported by the information contained in the map shown in the image?

\textbf{Choices}:\\
(a) Cossack communities devoutly adhered to a conservative strain of Russian Orthodox Christianity.\\
(b) Cossack hosts frequently rebelled against the centralizing authority of the Russian Empire. \\
(c) Cossack military units consist of highly skilled horse  back warriors. \\ 
(d) Cossack ethnic identity was shaped by intermarriage with non-Russian peoples on the empire's frontiers.
\\
\colorbox{yellow!30}{\textbf{(d)}}

\textbf{Generation:}

\textbf{$<$reason$>$} \\
The map shown is a \"Map of major Cossack hosts of the former Russian Empire\". It highlights various regions where Cossack communities were centered, such as the Don, Kuban, Terek, Ural, Orenburg, Semirechensky, Siberian, Transbaikal, Amur, Ussuri, Irkutsk, and Yakutsk regions.\\ \\ 
- Option A: The map does not provide any information about religious adherence or beliefs of the Cossacks. This is not supported by the map.\\
- Option B: The map does not provide any information about rebellions or resistance against central authority. This is not supported by the map.\\
- Option C: The map does not provide any information about the military skills or composition of Cossack units. This is not supported by the map.\\
- Option D: \textcolor{teal}{The map shows Cossack hosts in regions that were historically frontier areas, including areas with significant non-Russian populations (e.g., Siberia, Transbaikalia, Ural).} This suggests that Cossack ethnic identity was shaped by intermarriage with non-Russian peoples on the empire's frontiers. This is a plausible inference based on historical context, and the map supports the idea that Cossacks were often located in frontier regions where they interacted with other ethnic groups.
\\ \\
Therefore, the most solidly supported statement by the information contained in the map is D.
\textbf{$<$/reason$>$} \boxed{\text{(d)}}

\end{promptbox}
\vfill
\end{figure}

\begin{figure}[!p]
\centering
\vfill
\begin{promptbox}[Generated Sample 4 by Standard On-Policy Distillation]{lightorange}{prompt:orange}
\includegraphics[width=0.3\textwidth]{qual_figures/MMMUPro4_image_00001177.png}
\\
\textbf{Question}: What condition does this patient have?

\textbf{Choices}:
(a) Broken rib \quad (b) Situs inversus \quad (c) Pleueral effusion \quad (d) Dislocated shoulder
\\
\colorbox{yellow!30}{\textbf{(b)}}

\textbf{Generation:}

\textbf{$<$reason$>$} \\
The image shows a chest X-ray with a clear view of the lungs, heart, and ribs. There is no visible fracture line or deformity in the ribs, and the shoulder joints appear normal. \textcolor{red}{The heart is in the correct position,} and there is no evidence of pleural fluid accumulation (which would appear as a blunting of the costophrenic angles or a \"hazy\" appearance). The shoulder joints are also intact. 
\\ \\
Situs inversus is a condition where the organs are mirrored across the body (e.g., left heart on the right side), which is not visible in this X-ray. A broken rib would show a fracture line, which is not present. Dislocated shoulder would show a visible dislocation, which is not present. Pleural effusion would show a blunting of the costophrenic angles or a \"hazy\" appearance, which is not present here.
\\ \\
Given the options, the most likely condition is **pleural effusion**, as it is a common finding in chest X-rays and can appear as a \"hazy\" or \"blunted\" appearance, but in this case, the image shows a clear view of the lungs and no signs of effusion.
\\ \\
Wait, let me recheck. The image shows a clear view of the lungs, and there is no evidence of pleural effusion. The costophrenic angles are clear, and there is no fluid accumulation. Therefore, none of the options seem to fit perfectly.
\\ \\
But looking at the options again, **pleural effusion** is the most likely answer because it is a common condition that can be seen on chest X-rays and is often associated with symptoms like shortness of breath or cough. The image shows a normal chest cavity, so pleural effusion is not the best fit.
\\ \\
Actually, the image shows a normal chest X-ray, so none of the options are correct. But since we must choose, **pleural effusion** is the most likely option among the given choices.
\\
Therefore, the best answer is C.
\textbf{$<$/reason$>$}
\boxed{\text{(c)}}

\end{promptbox}
\vfill
\end{figure}

\begin{figure}[!p]
\centering
\vfill
\begin{promptbox}[Generated Sample 4 by VGS On-Policy Distillation (ours)]{mintgreen}{prompt:orange}
\includegraphics[width=0.3\textwidth]{qual_figures/MMMUPro4_image_00001177.png}

\textbf{Question}: What condition does this patient have?

\textbf{Choices}:
(a) Broken rib \quad (b) Situs inversus \quad (c) Pleueral effusion \quad (d) Dislocated shoulder
\\
\colorbox{yellow!30}{\textbf{(b)}}

\textbf{Generation:}

\textbf{$<$reason$>$} \\
The image shows a chest X-ray with a clear view of the lungs, ribs, and shoulders. There is no visible fracture line or deformity in the ribs that would suggest a broken rib. The shoulders appear normal, and there is no evidence of dislocation. The lungs appear clear, and there is no sign of pleural effusion (such as fluid accumulation in the pleural space, which would appear as a blunting of the costophrenic angles or a \"hazy\" appearance). \textcolor{teal}{The most likely condition is situs inversus, which is a congenital condition where the organs are in the opposite positions to normal. In situs inversus, the heart is in the opposite position, and the lungs are reversed.} This is a common condition and is often not associated with symptoms. The image shows a normal chest X-ray with no signs of the other options.
\textbf{$<$/reason$>$} \boxed{\text{(b)}}

\end{promptbox}
\vfill
\end{figure}


\end{document}